\documentclass[11pt]{article}

\usepackage[final]{acl}

\usepackage{times}
\usepackage{latexsym}

\usepackage[T1]{fontenc}

\usepackage[utf8]{inputenc}

\usepackage{microtype}

\usepackage{inconsolata}

\usepackage{graphicx}
\usepackage{booktabs}
\usepackage{amsmath} 
\usepackage{algorithm}
\usepackage{algpseudocode}
\usepackage{caption}
\usepackage{enumitem}
\usepackage{subcaption}
\usepackage{setspace}
\usepackage[table]{xcolor}
\usepackage{makecell}
\usepackage{array}
\usepackage{wrapfig}
\usepackage{multirow}

\usepackage[accsupp]{axessibility}  
\usepackage{nicefrac}
\usepackage{natbib} 
\usepackage{marvosym}

\title{Shallow to Deep: Aligning Token Pruning with Stage-wise Roles in LVLMs}

\author{
    \textbf{Shuo Zhang}\thanks{Equal contribution.} \quad
    \textbf{Jintao Tong}\footnotemark[1] \quad
    \textbf{Yixiong Zou}\textsuperscript{\Letter} \quad
    \textbf{Yuhua Li} \quad
    \textbf{Ruixuan Li} \\
    School of Computer Science and Technology, Huazhong University of Science and Technology \\
    \texttt{\{zhangshuo, jintaotong, yixiongz, idcliyuhua, rxli\}@hust.edu.cn}
}

\begin{document}
\maketitle
\begingroup
\renewcommand{\thefootnote}{\Letter}
\footnotetext{Corresponding author.}
\endgroup

\renewcommand{\multirowsetup}{\centering}
\definecolor{mygray}{gray}{.92}
\definecolor{ForestGreen}{RGB}{34,139,34}
\newcommand{\fg}[1]{\mathbf{\textcolor{ForestGreen}{#1}}}
\definecolor{Forestred}{RGB}{220,50,50}
\definecolor{lightgreen}{rgb}{0.886, 0.941, 0.851}
\newcommand{\fr}[1]{\mathbf{\textcolor{Forestred}{#1}}}

\begin{abstract}
 Large Vision-Language Models (LVLMs) incur high computational costs from redundant visual tokens. Although training-free attention-based multi-layer pruning in the vision encoder stage has been explored as an effective strategy, we find that pruning in shallow layers consistently degrades performance. In this paper, we aim to understand this problem and seek a solution. By analyzing attention patterns across network depth, we find that shallow layers primarily function as edge detectors with chaotic attention maps, while deeper layers transition through local subject recognition and unstable semantic aggregation. To address the misalignment between pruning strategies and network stages, we propose \textit{STD}, a hierarchical token pruning framework that adapts token selection mechanisms to the functional role of each network stage. STD employs High-Frequency Spectral Analysis in shallow layers to deterministically preserve structural edges, uses Gaussian-Smoothed Attention in intermediate layers to maintain spatial coherence, and introduces a Stability-Adaptive Trigger in deep layers to execute pruning only during semantically stable phases. Extensive experiments show that STD outperforms state-of-the-art pruning methods by 1.1\% on LLaVA-1.5-7B with 88.9\% token reduction, while also being plug-and-play and highly effective when combined with other methods, and by 2.1\% on LLaVA-NeXT-7B with 94.4\% reduction, delivering a 3.9× speed-up in the prefilling stage. Our code will be released at \url{https://github.com/Twilight03/STD}.
\end{abstract}

\begin{figure}[t]
    \centering
    
    \begin{minipage}{0.38\textwidth}
        \centering
        \includegraphics[width=\textwidth]{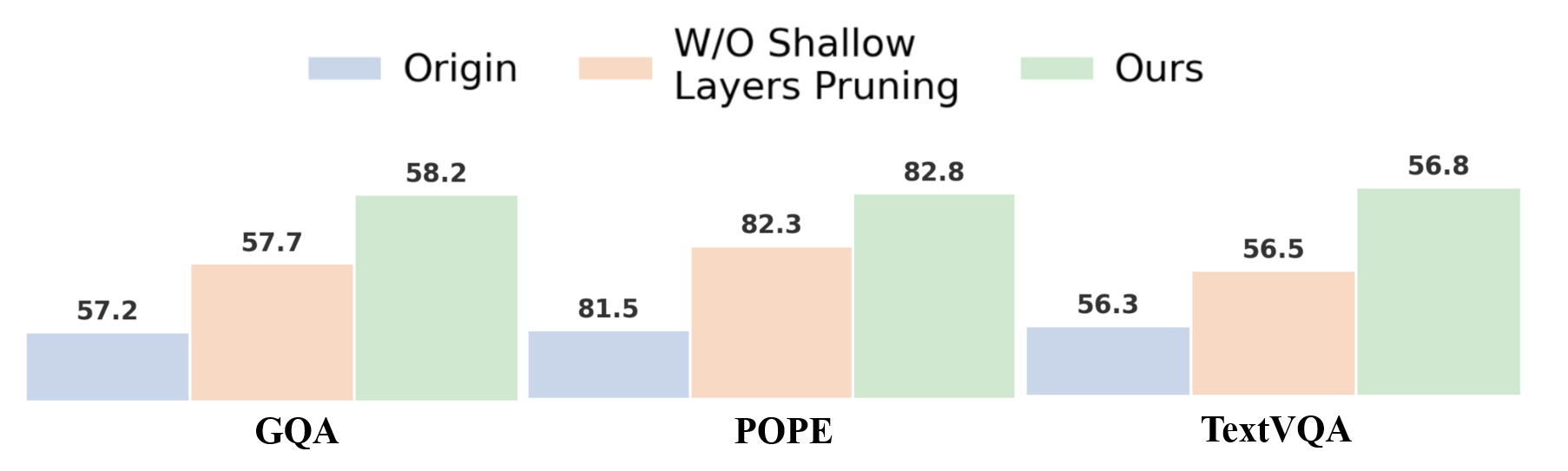}
    \end{minipage}
    
    \begin{minipage}{0.38\textwidth}
        \centering
        \includegraphics[width=\textwidth]{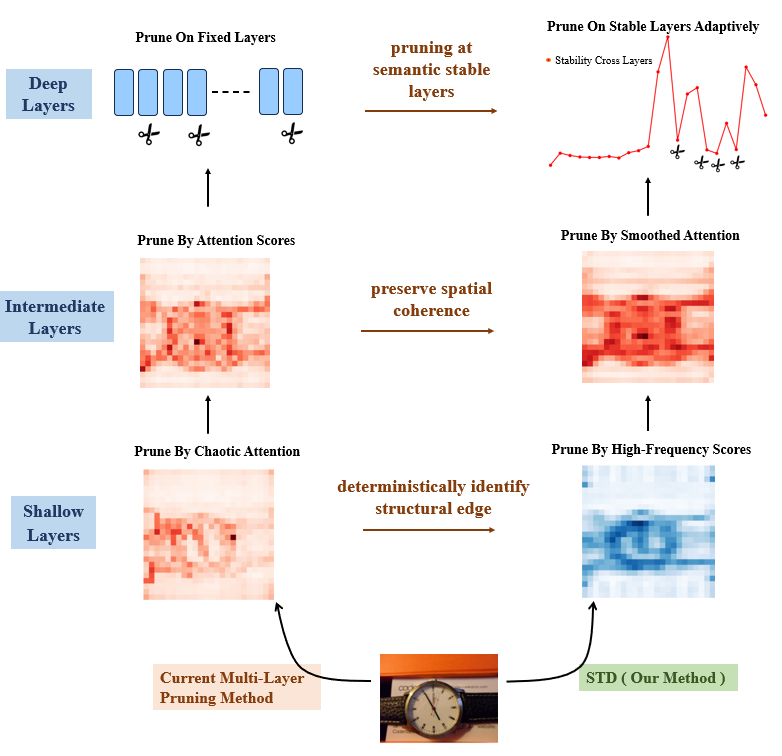}
    \end{minipage}
    \vspace{-0.2cm}
    \caption{\textbf{(Top)} Current works suffer from the degraded performance in pruning shallow-layer visual tokens, which is under-explored so far. \textbf{(Bottom)} In this paper, we handle this problem by revisiting stage-wise roles of the vision encoder, and design a method \textbf{STD} to align with these roles, improving the maintained semantics of each stage.}
    \label{fig:combined_overview}
    \vspace{-0.3cm}
\end{figure}

\vspace{-0.2cm}
\section{Introduction}
Large Vision-Language Models (LVLMs)~\cite{bai2023qwenvl,li2023blip,liu2023visual,tong2024cambrian} incur high cost when processing massive visual token sequences, especially for high-resolution inputs~\cite{chen2024image,liu2024improved,wang2024qwen2}. To improve efficiency, training-free visual token pruning methods have been proposed. 
Compared with pruning at the LLM stage using cross-modal attention~\cite{zhang2024sparsevlm,xing2024pyramiddrop,wu2026hidrop}, pruning in the vision encoder or projection layer is more efficient because redundant tokens are removed before entering the LLM~\cite{yang2024visionzip,tong2025flowcut,han2026filter,chen2026evoprune}. Among these methods, attention-based multi-layer pruning has become an effective approach~\cite{han2026filter,tong2025flowcut,chen2026evoprune}. It progressively discards low-attention tokens from shallow to deep layers, reducing redundancy and accelerating inference while preserving performance.

Despite the success of multi-layer strategies, we find that standard pruning in shallow layers often harms the model, making pruning in these layers ineffective. As illustrated in \textbf{Fig.~\ref{fig:combined_overview}} (Top), disabling pruning in shallow layers while retaining it in deeper stages yields better performance than standard full-layer pruning. Although this issue is common in existing methods, to the best of our knowledge, it has rarely been studied.

In this paper, we aim to understand this problem and seek a solution. Since attention is a widely used pruning criterion~\cite{zhang2024sparsevlm,yang2024visionzip,tong2025flowcut}, we analyze attention patterns across network depths, especially in shallow layers. Existing works~\cite{park2022vision} generally assume token interactions with neighbors in shallow layers.. However, we find that this assumption does not hold in the very shallow layers. Instead, their attention maps are inherently \textbf{chaotic and noisy}, failing to capture meaningful semantic structures and exhibiting random information flow. Therefore, using these chaotic signals can mistakenly remove critical information.

However, this chaos in the very shallow layers is not merely noise. We further show that \textbf{the model operates primarily as an edge detector when semantic understanding is immature}. At this stage, the network focuses on high-frequency contours rather than objects, indicating that token importance is more related to spectral properties than to attention weights. As the network deepens, it shifts from frequency-based edge perception to recognition of local subjects, requiring spatial coherence for object integrity. Finally, the model enters an abstract semantic aggregation phase: distant image regions exchange information through hub tokens, resembling the attention-sink phenomenon~\cite{yi2026attention}, and gradually converge into abstract whole-image semantic representations, where we observe \textbf{instability} in semantic evolution.

Based on the above observations, we propose \textbf{STD}, a hierarchical token pruning framework that aligns pruning strategies with the functional roles of different network stages. As shown in \textbf{Fig.~\ref{fig:combined_overview}} (Right), STD redesigns pruning in three phases: (1) In shallow layers, we replace noisy attention weights with \textit{High-Frequency Spectral Analysis} to deterministically identify structural edge features; (2) In intermediate layers, we use \textit{Gaussian-Smoothed Attention} to preserve spatial coherence and feature continuity; (3) In deep layers, we introduce a \textit{Stability-Adaptive Trigger} that prunes only at semantically stable moments.
Extensive experiments show that STD is effective on image and video understanding tasks, remains plug-and-play and compatible with other methods, and substantially improves inference efficiency.

In summary, our main contributions are: 

\textbullet\ We identify the failure mechanism of conventional shallow layer pruning, attributing it to chaotic attention patterns misaligned with the edge-detection role of early layers.
\begin{figure*}[t]
    \centering
    \includegraphics[width=0.92\linewidth]{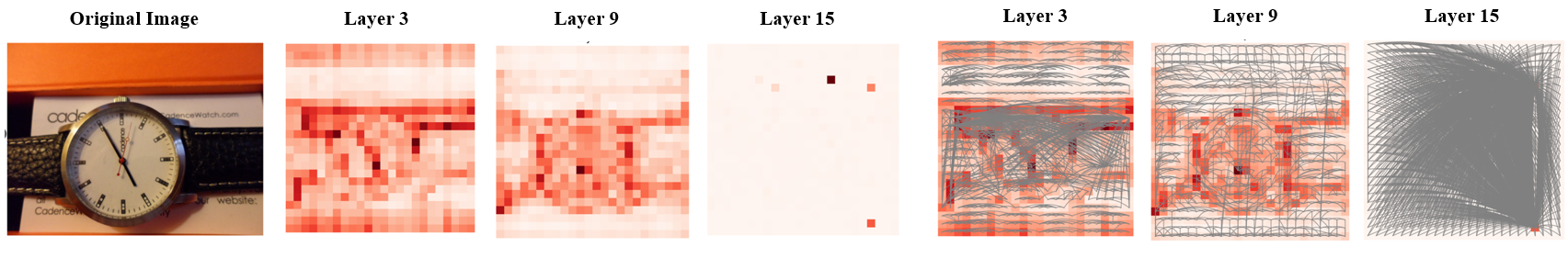}
    \vspace{-0.08cm}
    \caption{(Left) CLS token attention heatmaps at layers 3, 9 and 15, showing \textbf{chaotic and noisy} distributions in shallow layers (3), focused on main objects in middle layers (9), and hub token concentration in deep layers (15). (Right) Information flow visualization of connections between tokens and their top-3 attended tokens, showing irregular flow in shallow layers, stable local structures in middle layers, and hub-centric patterns in deep layers.}
    \label{fig:attention_chaotic}
    \vspace{-0.05cm}
\end{figure*}
\begin{figure*}[t]
    \centering
    \begin{minipage}{0.30\textwidth}
        \centering
        \includegraphics[width=\textwidth]{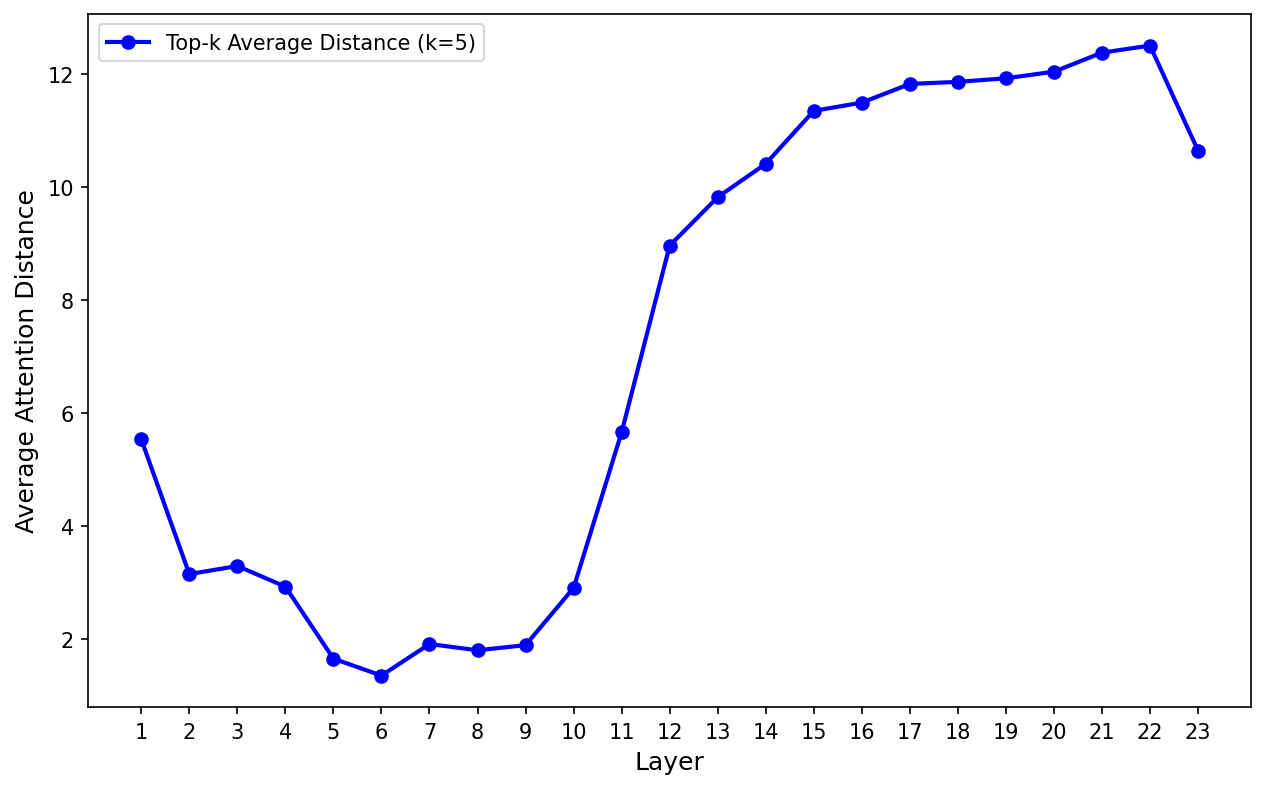}
    \end{minipage}
    \hfill
    \begin{minipage}{0.30\textwidth}
        \centering
        \includegraphics[width=\textwidth]{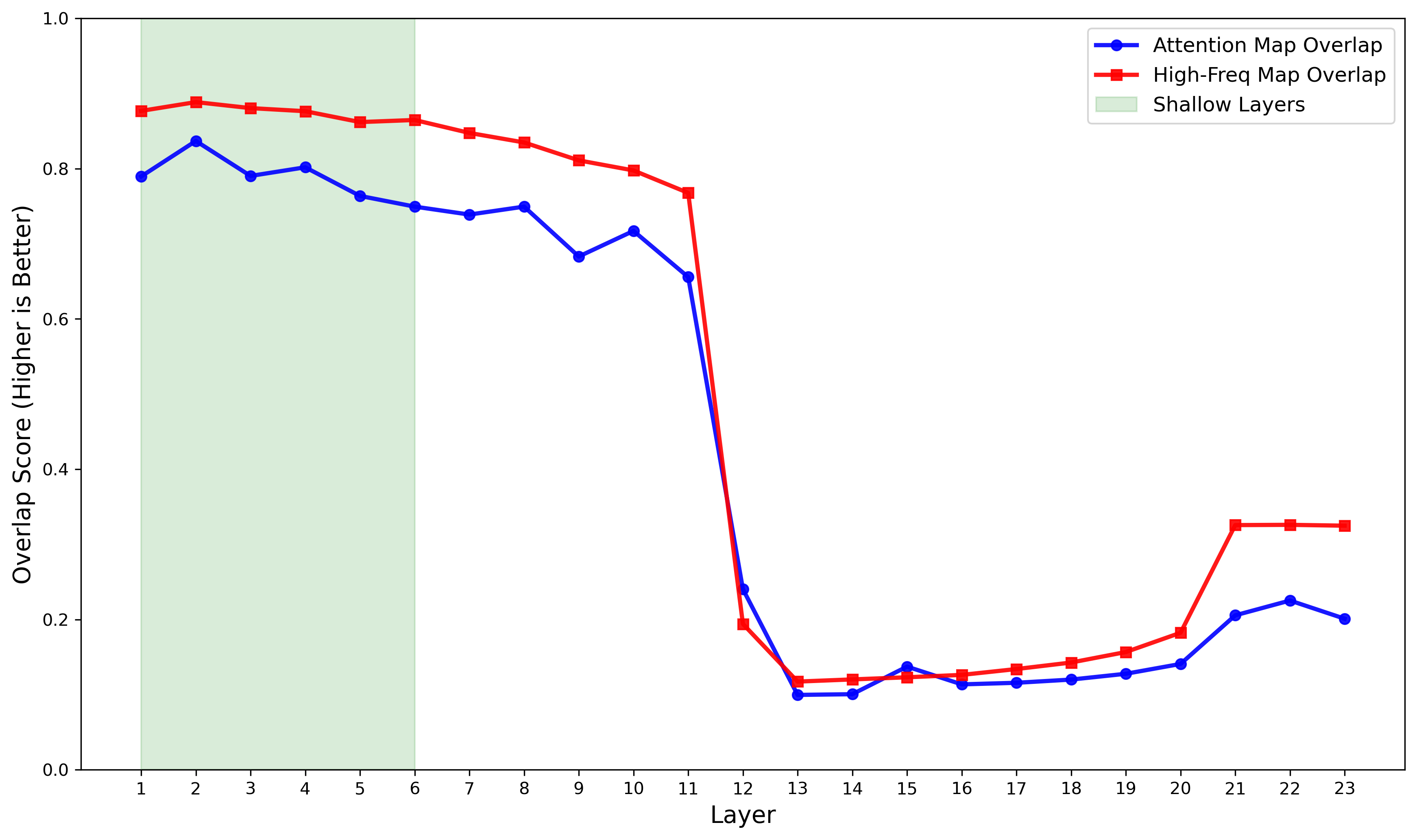}
    \end{minipage}
    \hfill
    \begin{minipage}{0.30\textwidth}
        \centering
        \includegraphics[width=\textwidth]{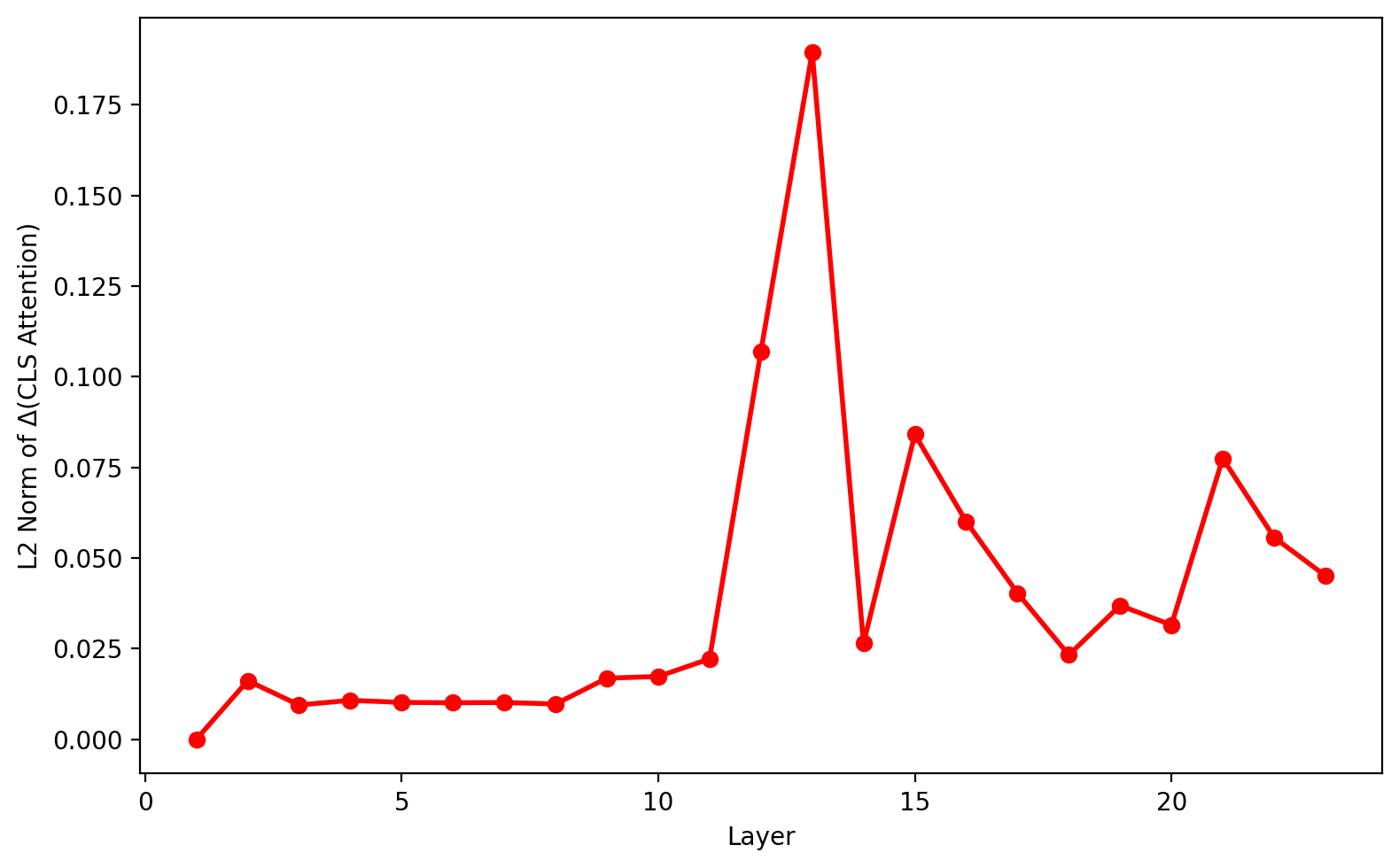}
    \end{minipage}
    
    \vspace{-0.08cm}
    \caption{
    (Left) Average attention distance exhibits a ``U-shaped'' curve, challenging the view that shallow layers only attend to neighbors;
(Middle) Edge detection accuracy shows high-frequency analysis is a more deterministic proxy for token selection than attention weights;
(Right) Semantic change instability in the deep stage alternates between rapid semantic evolution and relative stability.}
    \label{fig:combined_attn_analysis}
    \vspace{-0.5cm}
\end{figure*}

\textbullet\ We characterize LVLM functional evolution across depth: shallow layers detect high-frequency structural contours, intermediate layers shift to local subject recognition, and deep layers perform abstract semantic aggregation where information converges through inherently unstable hub tokens;

\textbullet\ We propose STD, a unified approach that adapts pruning to each stage's role: shallow layers use high-frequency spectral analysis for deterministic edge preservation, intermediate layers use Gaussian-smoothed attention for spatial coherence, and deep layers use stability-adaptive triggering for stable semantic aggregation;

\textbullet\ Experiments show that STD outperforms SoTA by 1.1\% on LLaVA-1.5-7B with 88.9\% token reduction, and gains a further 1.8\% when combined with other methods. On LLaVA-NeXT-7B, STD improves SoTA by 2.1\% with 94.4\% token reduction and achieves a 3.9$\times$ prefilling speed-up.

\section{Rethinking Model's Stage-wise Roles from Shallow-layer Failure}




\subsection{Shallow-layer Pruning Harms Accuracy}
\vspace{-0.15cm}

\begin{table}[t]
    \centering
    \vspace{-0.2cm}
    \caption{\footnotesize{Performance comparison of attention-based multi-layer pruning with and without shallow layer pruning. Results reveal the flaw of indiscriminate pruning in early stages.}}
    \vspace{0cm}
    \label{tab:shallow_pruning}
    
    \footnotesize 
    \setlength{\tabcolsep}{3pt}
    \renewcommand{\arraystretch}{1.05}
    \resizebox{0.9\columnwidth}{!}{
    \begin{tabular}{@{} l | *{4}{c} | c @{}}
        \toprule[1.2pt]
        \textbf{Method} 
        & \textbf{GQA} 
        & \textbf{MME} 
        & \textbf{POPE} 
        & \textbf{TextVQA} 
        & \textbf{Avg}. \\
        \hline
        
        \rowcolor{mygray}
        \multicolumn{6}{c}{\textit{Retain 192 Tokens}}\\
        Full Multi-layer Pruning & 58.3 & 1716 & 84.3 & 56.7 & 95.4\% \\
        \textbf{w/o Shallow Pruning} & \textbf{58.6} & \textbf{1742} & \textbf{84.9} & \textbf{56.9} & \textbf{96.1\%} \\
        \hline
        
        \rowcolor{mygray}
        \multicolumn{6}{c}{\textit{Retain 128 Tokens}}\\
        Full Multi-layer Pruning & 57.2 & 1669 & 81.5 & 56.3 & 93.3\% \\
        \textbf{w/o Shallow Pruning} & \textbf{57.7} & \textbf{1693} & \textbf{82.3} & \textbf{56.5} & \textbf{94.1\%} \\
        \hline
        
        \rowcolor{mygray}
        \multicolumn{6}{c}{\textit{Retain 64 Tokens}}\\
        Full Multi-layer Pruning & 54.7 & 1607 & 79.2 & 54.5 & 90.5\% \\
        \textbf{w/o Shallow Pruning} & \textbf{55.0} & \textbf{1646} & \textbf{79.8} & \textbf{54.7} & \textbf{91.0\%} \\
        \bottomrule[1.2pt]
    \end{tabular}}
    \vspace{-0.1cm}
\end{table}

Attention-based pruning is widely used for visual token compression in the Vision Encoder Stage, estimating token importance via attention weights. Compared with \textit{single-layer} pruning, which removes tokens once at a selected layer, \textit{multi-layer} pruning is generally more effective. As token redundancy emerges progressively as the network deepens; removing redundant tokens early helps avoid ineffective dispersion of attention resources.

\begin{figure*}
\centering
\includegraphics[width=0.90\linewidth]{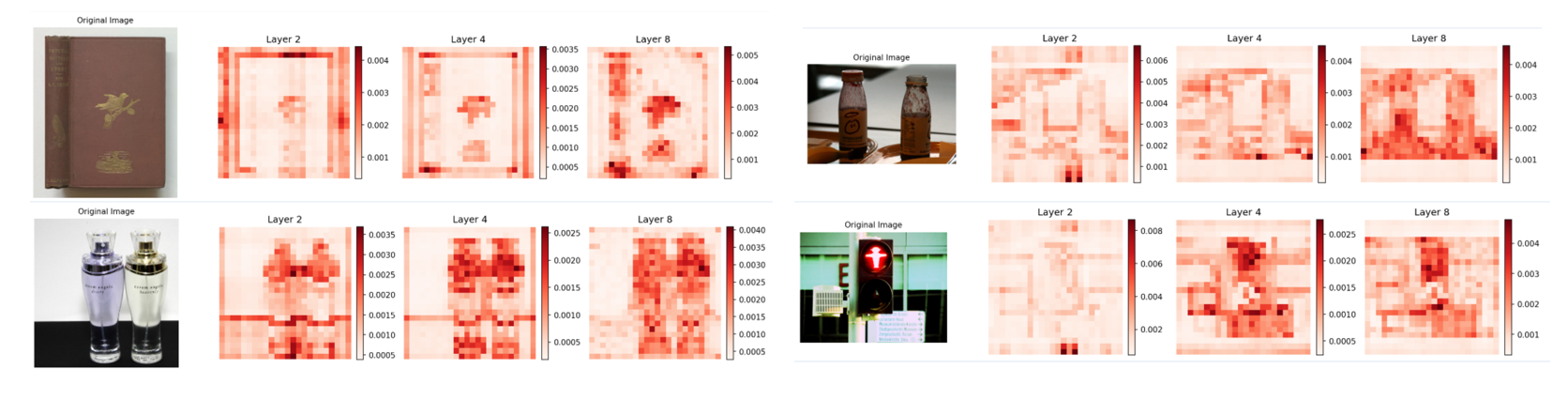}
\vspace{-0.08cm}
\caption{Attention distribution patterns in shallow and middle layers for three samples. Shallow layers (2, 4) primarily focus on \textbf{edge contours}, while middle layers (8) successfully identify complete semantic subjects.}
\label{fig:attn_edge}
\vspace{-0.4cm}
\end{figure*}

Although attention-based multi-layer pruning outperforms single-layer approaches, we find a counter-intuitive phenomenon: \textbf{indiscriminate attention-based pruning in shallow layers degrades model performance.}
To verify this, we conduct a controlled study on four benchmarks under three token retention budgets. For each budget, we compare (1) Standard Multi-layer Pruning, applying attention-based pruning every two layers, and (2) w/o Shallow Pruning, disabling pruning in shallow layers while keeping it in middle and deep layers. The final pruning step strictly enforces the token retention budget.
As shown in Table~\ref{tab:shallow_pruning}, the latter consistently outperforms the former across all benchmarks and budgets.

This finding shows that shallow-layer attention weights are unreliable signals. Existing attention-based multi-layer pruning assumes that attention is equally informative across all depths, but this fails in early layers. Motivated by this, we analyze why attention-based pruning fails in shallow layers.


\vspace{-0.22cm}
\subsection{Attention Fails in Shallow-Layer Pruning}

\textbf{Chaotic Shallow-Layer Attention.}
To explain the poor performance of shallow-layer pruning, we analyze CLS token attention patterns. Fig.~\ref{fig:attention_chaotic} (Left) shows that shallow layers (3) have \textbf{chaotic and noisy} distributions, often attending to padding and edges rather than semantic objects. In contrast, middle layers (9) consistently focus on objects, while deep layers (15) converge on hub tokens.

\textbf{Disordered Information Flow.}
Existing analyses of information flow~\cite{tong2025flowcut} often assume that shallow layers mainly aggregate local neighborhood information. However, as shown in Fig.~\ref{fig:attention_chaotic} (right), the very shallow layers instead show irregular attention patterns: tokens attend not only to nearby neighbors but also to distant irrelevant positions, and even low-semantic tokens can become hubs. The local structures attributed to shallow layers in prior work become clear only in the deeper shallow layers (i.e., the middle layers), while deep layers are dominated by hub-centric patterns.

\textbf{Abnormal Attention Distances.}
Conventional wisdom suggests that shallow layers mainly attend to nearby tokens, resulting in low attention distances~\cite{dosovitskiy2020image,park2022vision,liu2022convnet}. However, Fig.~\ref{fig:combined_attn_analysis} (right) reveals a U-shaped trend: attention distances are unexpectedly high in the very shallow layers due to noisy patterns, decrease in the deeper shallow layers (i.e., the middle layers) as local object recognition becomes more coherent, and rise again in the deep layers as attention converges onto hubs.

\textbf{Summary.}
The observation of shallow layers made by prior works are mainly about the ``deeper shallow layer'', but not for the ``very shallow layer'', leading to misalignment between pruning criterion and layer roles and finally harms performance.

\vspace{-0.2cm}
\subsection{Stage-wise Evolution of Functional Roles}
\label{sec:stage_wise_evol}
To analyze the very shallow layers, we divide the shallow layer'' into the very shallow layer'' and ``deeper shallow layer'', hereafter denoted as \textit{shallow layer} and \textit{middle/intermediate layers}.    

\noindent\textbf{(a) Model Perception Stage: From Edge Detection to Object-Level Semantic.}     

Although shallow layers appear semantically chaotic, they are not noise but play a distinct role in pattern learning. As shown in Figure \ref{fig:attn_edge}, in the shallow stage, the model focuses on object edge contours rather than semantic subjects, suggesting that \textbf{the model operates primarily as an edge detector when semantic understanding is immature}. Given this, we hypothesize that frequency-domain analysis offers a more deterministic and accurate proxy for token selection. Since high-frequency components capture edges and fine details~\cite{tong2024lightweight}, they provide a more reliable way to identify edge-related tokens than attention patterns.
\begin{figure*}[t]
    \centering
    \includegraphics[width=0.87\linewidth]{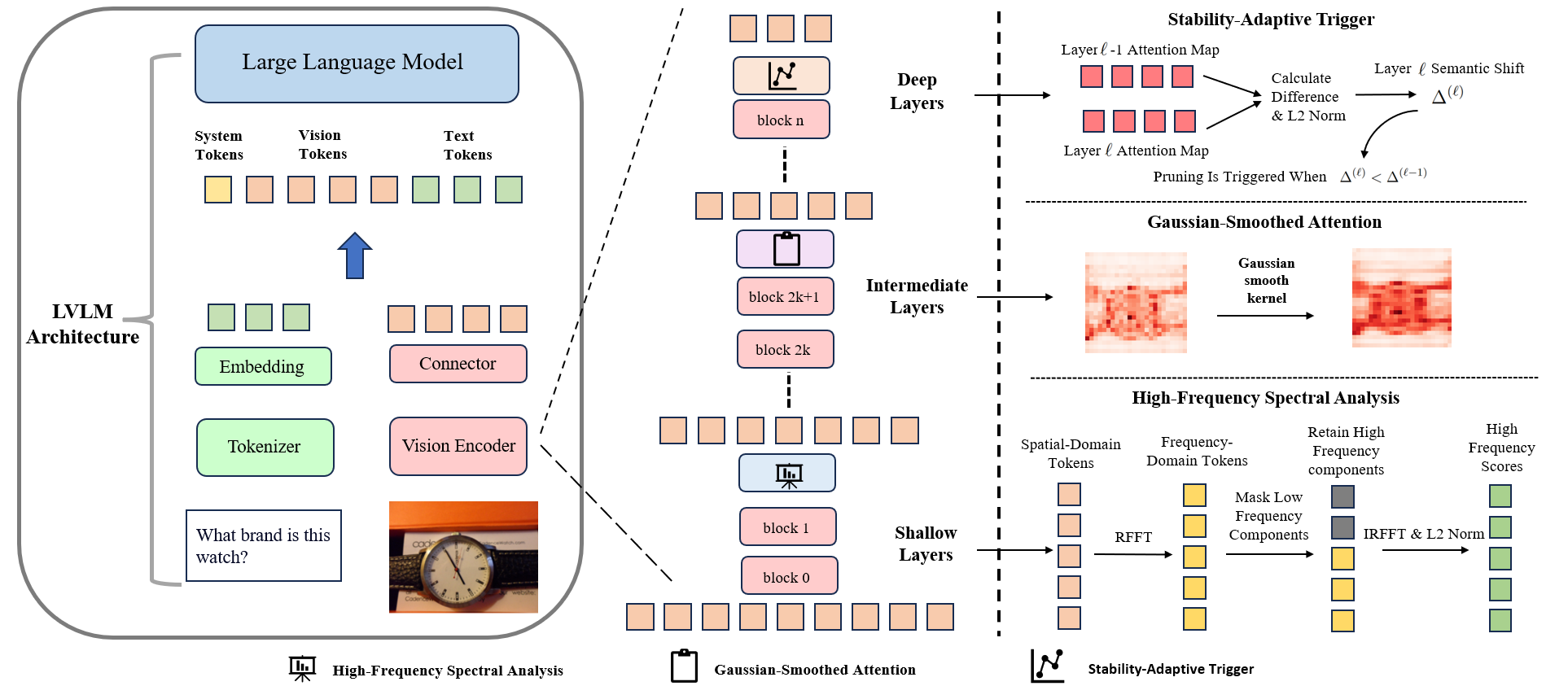}
    \caption{\textbf{Overview of the STD Framework.} 
    The pipeline adapts pruning strategies across network depths, employing stage-specific strategies to replace the original uniform attention-based approach.}
    \label{fig:method_overview}
    \vspace{-0.5cm}
\end{figure*}

\vspace{-0.4cm}
To quantitatively verify the role of shallow layers in edge detection, Fig.~\ref{fig:combined_attn_analysis} (middle) compares tokens selected via Canny edge detection, attention scores, and high-frequency analysis. The results show high similarity between edge maps and attention weights in shallow layers ($\approx 0.8$), which decreases with depth, supporting the edge detector hypothesis. Moreover, across shallow layers (1--6), high-frequency scores consistently show higher similarity to the Canny reference than attention-based scores. This confirms that high-frequency spectral analysis provides more accurate edge localization than erratic attention patterns.

The above observation suggests a plausible empirical pattern: shallow layers tend to detect edge contours, while intermediate layers progressively capture object-level features. This edge-to-object progression is consistent with the hierarchical learning of visual representations.


\noindent\textbf{(b) Abstract Semantic Aggregation Stage: Instability of Semantic Integration}

Deep layers process information differently from shallow and middle layers. Beyond a certain layer, attention rapidly concentrates on a few hub tokens, marking the \textbf{abstract semantic aggregation phase}. In this phase, token interaction is mainly mediated by hub tokens, and the model shifts from perceiving diverse visual content to aggregating abstract semantics. Given this change in processing mode, local subject retention no longer applies, so we instead emphasize the stability of inter-layer semantic changes.

The semantic shifts in deep layers are not uniform but show significant \textbf{instability}. In Fig.~\ref{fig:combined_attn_analysis}(Right), we compute the difference between adjacent layers ($\Delta$ CLS Attention) and use the $L_2$ norm to measure its magnitude. During the perception stage, this change remains low and stable; during the deep-layer semantic-aggregation phase, it fluctuates significantly. At layers with a high $L_2$ norm of differences (i.e., unstable layers), the pattern evolves effectively between adjacent layers, indicating that the model extracts useful patterns from these layers, so pruning tokens there is harmful. In contrast, at layers with a low $L_2$ norm of differences (i.e., stable layers), adjacent-layer patterns are similar, implying less information is extracted and more redundancy may exist. Consequently, pruning at these layers will be harmless.

\textbf{In summary}, our empirical observations suggest a functional pattern across shallow to deep layers: shallow layers respond to to high-frequency edges, middle layers exhibit stronger Object-Level semantic recognition, and deep layers tend to aggregate semantic information while showing greater instability. In the following sections, we propose a method designed to better align with these roles.


\vspace{-0.2cm}
\section{Methodology}
\vspace{-0.2cm}

Building upon our analysis of visual attention evolution, we propose \textbf{STD}, a \textbf{hierarchical token pruning framework} aligned with the functional roles of different network stages. As shown in \textbf{Fig.~\ref{fig:method_overview}}, STD uses three phases: (1) replacing noisy attention weights with \textit{High-Frequency Spectral Analysis} in shallow layers; (2) using \textit{Gaussian-Smoothed Attention} in intermediate layers; and (3) applying \textit{Stability-Adaptive Trigger} in deep layers.


\subsection{Phase 1: Shallow Layers via High-Frequency Spectral Analysis}
\label{subsec:phase1_shallow}

In shallow layers ($\ell \leq L_1$), attention patterns are \textbf{chaotic and noisy} and thus unreliable for token selection. We therefore replace them with \textit{High-Frequency Spectral Analysis} to \textbf{deterministically identify structural edge features}. 

Let $\mathbf{P}^{(\ell)} = [\mathbf{p}_1^{(\ell)}, \dots, \mathbf{p}_N^{(\ell)}]^\top \in \mathbb{R}^{N \times D}$ denote the sequence of $N$ patch tokens at layer $\ell$. We apply the real-valued fast Fourier transform (RFFT) along the token sequence:
\begin{equation}
    \widehat{\mathbf{P}}^{(\ell)} = \mathcal{F}_{\text{rfft}}\left( \mathbf{P}^{(\ell)} \right) \in \mathbb{C}^{F \times D},
\end{equation}
where $F = \lfloor N/2 \rfloor + 1$. To isolate high-frequency information corresponding to edge features, we retain the upper two-thirds of the spectrum:
\begin{equation}
    \widehat{\mathbf{P}}^{(\ell)}_{\text{high}}[f, :] = 
    \begin{cases}
        \widehat{\mathbf{P}}^{(\ell)}[f, :], & f \geq \lceil F/3 \rceil \\
        0, & \text{otherwise}
    \end{cases}.
\end{equation}
The high-frequency representation is reconstructed by inverse RFFT: $\widetilde{\mathbf{P}}^{(\ell)}_{\text{high}} = \mathcal{F}_{\text{irfft}}(\widehat{\mathbf{P}}^{(\ell)}_{\text{high}}; n=N)$. The high-frequency energy score of token $i$ is:
\begin{equation}
    s_i^{\text{high}, (\ell)} = \left\| \widetilde{\mathbf{p}}_i^{(\ell), \text{high}} \right\|_2^2.
\end{equation}

To ensure a smooth transition, we combine the high-frequency score with attention weights:
\begin{equation}
    r_i^{(\ell)} = \alpha \cdot s_i^{\text{high}, (\ell)} + (1 - \alpha) \cdot a_i^{(\ell)}, \quad \ell \leq L_1,
\end{equation}
where $\alpha$ balances frequency cues and attention, and $a_i^{(\ell)}$ is the attention weight from the [CLS] token to patch token $i$ at layer $\ell$. All scores are normalized to $[0,1]$ by min-max scaling. 

\subsection{Phase 2: Intermediate Layers via Gaussian-Smoothed Attention}
\label{subsec:phase2_intermediate}

In intermediate layers ($L_1 < \ell \leq L_2$), the model begins subject-level semantic recognition, and attention becomes more reliable with strong locality around main subjects. We therefore use attention weights with \textit{Gaussian-Smoothed Attention} to \textbf{preserve spatial coherence and ensure robust feature continuity}. This reduces object-part fragmentation caused by hard thresholding.

The smoothed attention score is:
\begin{equation}
    r_i^{(\ell)} = \sum_{\delta = -\lfloor K/2 \rfloor}^{\lfloor K/2 \rfloor} \mathcal{G}_\sigma(\delta) \cdot a_{i+\delta}^{(\ell)}, \quad L_1 < \ell \leq L_2,
\end{equation}
where $\mathcal{G}_\sigma$ is a 1D Gaussian kernel with standard deviation $\sigma$, and $K$ is the kernel size. 

For shallow and intermediate layers, pruning is performed \textbf{every two layers} to progressively reduce redundancy as the model builds semantic awareness in the perception stage.


\begin{table*}[t]
\centering
\caption{\textbf{Performance evaluation of STD on LLaVA-1.5-7B under varying configurations.} With 576 visual tokens as the baseline, the final column denotes the average accuracy normalized to the theoretical upper bound.}
\vspace{-0.05cm}
\label{tab:main}
\setstretch{0.86}
\small
\setlength{\tabcolsep}{2pt}
\resizebox{0.85\linewidth}{!}{
\begin{tabular}{l | c c c c c c c c | >{\centering\arraybackslash}p{1.2cm}}
\toprule[1.2pt]
\small\textbf{Method} & \small\textbf{ GQA } & \small\textbf{MMB} & \small\textbf{MMB}$^{\text{CN}}$ & \small\textbf{MME} & \small\textbf{ POPE } & \small\textbf{ SQA } & \small\textbf{VQA}$^{\text{V2}}$ & \small\textbf{VQA}$^{\text{Text}}$ & \makecell[c]{\textbf{Avg}.}\\
\hline

\rowcolor{mygray}
\multicolumn{9}{c}{\textit{Upper Bound, 576 Tokens} \ $\textbf{(100\%)}$}\\
\textcolor{gray}{Vanilla} & \textcolor{gray}{61.9} & \textcolor{gray}{64.7} & \textcolor{gray}{58.1} & \textcolor{gray}{1865} & \textcolor{gray}{85.9} & \textcolor{gray}{69.5} & \textcolor{gray}{78.5} & \textcolor{gray}{58.3} & \multirow{1}*{\textcolor{gray}{100\%}} \\
\hline

\rowcolor{mygray}
\multicolumn{9}{c}{\textit{Retain 192 Tokens} \ $\fg{(\downarrow 66.7\%)}$}\\

PDrop \texttt{\scriptsize{(CVPR25)}} & 57.3 & 62.9 & 56.8 & 1766 & 82.3 & 68.8 & 75.1  & 56.5 & 96.2\% \\
SparseVLM \texttt{\scriptsize{(ICML25)}} & 57.6 & 62.5 & 53.7 & 1721 & 83.6 & 68.9 & 75.6 & 56.1 & 95.4\% \\
Flowcut \texttt{\scriptsize{(NeurIPS25)}} & 59.5 & 63.0 & 56.8 & 1833 & 85.7 & 68.4 & 76.8 & 57.3 & 97.8\% \\
FiCoCo-V \texttt{\scriptsize{(AAAI26)}} &58.5 &62.3 & 55.3 & 1732 & 82.5 & 67.8 & 74.4 & 55.7 & 95.3\%\\
V$^2$Drop \texttt{\scriptsize{(CVPR26)}} & 58.5 & \textbf{63.7} & 56.8 & 1826 & 85.1 &\textbf{69.3} & 75.2 & 55.6&97.3\% \\
\rowcolor{blue!10}
\textbf{STD (Ours)} & 60.1 & 63.1 & 57.4 & 1812 & \textbf{86.3} & 68.5 & 77.5 & 57.7 & 98.5\% \\
\rowcolor{blue!10}
\textbf{Flowcut + STD (Ours)}  & 60.3 & 63.5 & 57.4 & \textbf{1845} & \textbf{86.3} & 68.6 & \textbf{77.7} & 57.9 & 98.9\% \\
\rowcolor{blue!10}
\textbf{SparseVLM + STD (Ours)}  & \textbf{60.6} & \textbf{63.7} & \textbf{57.6} & 1818 & 86.2 & 68.9 & \textbf{77.7} & \textbf{58.0} & \textbf{99.0\%} \\
\hline

\rowcolor{mygray}
\multicolumn{9}{c}{\textit{Retain 128 Tokens} \ $\fg{(\downarrow 77.8\%)}$}\\
PDrop \texttt{\scriptsize{(CVPR25)}} & 57.1 & 61.6 & 56.3 & 1664 & 82.3 & 68.3 & 72.9 & 56.6 & 94.7\% \\
SparseVLM \texttt{\scriptsize{(ICML25)}} & 56.0 & 60.0 & 51.1 & 1696 & 80.5 & 67.1 & 73.8 & 54.9 & 92.5\% \\
Flowcut \texttt{\scriptsize{(NeurIPS25)}} & 58.2 & 61.7 & 56.0 & 1781 & 83.7 & 68.4 & 75.8 & 57.0 & 96.1\% \\
FiCoCo-V \texttt{\scriptsize{(AAAI26)}} &57.6 &61.1 &54.3 &1711 &82.2 &68.3 &73.1 &55.6& 94.3\%\\
V$^2$Drop \texttt{\scriptsize{(CVPR26)}} & 56.3& 61.8& 55.6&1712 &80.9 &\textbf{68.8} &73.7 &53.8 & 94.1\%\\
\rowcolor{blue!10}
\textbf{STD (Ours)} & 58.8 & 62.4 & \textbf{56.3} & 1763 & 84.6 & \textbf{68.8} & 76.2 & 57.4 & 97.0\% \\
\rowcolor{blue!10}
\textbf{Flowcut + STD (Ours)} & 59.1 & 62.5 & 56.1 & \textbf{1792} & 85.6 & 68.3 & 76.4 & 57.3 & 97.2\% \\
\rowcolor{blue!10}
\textbf{SparseVLM + STD (Ours)} & \textbf{59.6} & \textbf{62.9} & \textbf{56.3} & 1779 & \textbf{85.8} & 68.7 & \textbf{76.8} & \textbf{57.6} & \textbf{97.6\%} \\
\hline

\rowcolor{mygray}
\multicolumn{9}{c}{\textit{Retain 64 Tokens} \ $\fg{(\downarrow 88.9\%)}$}\\
PDrop \texttt{\scriptsize{(CVPR25)}} & 47.5 & 58.8 & 50.5 & 1561 & 55.9 & 68.6 & 69.2 & 50.6 & 84.6\% \\
SparseVLM \texttt{\scriptsize{(ICML25)}} & 52.7 & 56.2 & 46.1 & 1505 & 75.1 & 62.2 & 68.2 & 51.8 & 85.6\% \\
Flowcut \texttt{\scriptsize{(NeurIPS25)}} & 55.3 & 60.6 & 54.7 & 1712 & 79.3 & 68.9 & 72.8 & 55.6 & 93.2\% \\
FiCoCo-V \texttt{\scriptsize{(AAAI26)}} & 52.4&60.3 &53.0 &1591 &76.0 &68.1 &71.3 &53.6 & 90.5\%\\
V$^2$Drop \texttt{\scriptsize{(CVPR26)}} & 50.5& 55.2&53.1 &1470 &75.1 & 68.9&68.9 &51.8 & 87.5\%\\
\rowcolor{blue!10}
\textbf{STD (Ours)} & 56.1 & 61.1 & 55.3 & 1709 & 81.2 & 68.6 & 73.3 & \textbf{56.0} & 94.3\% \\
\rowcolor{blue!10}
\textbf{Flowcut + STD (Ours)} & \textbf{56.4} & \textbf{61.3} & \textbf{55.9} & \textbf{1744} & \textbf{81.6} & 68.8 & \textbf{73.9} & 55.9 & \textbf{95.0\%} \\
\rowcolor{blue!10}
\textbf{SparseVLM + STD (Ours)} & 54.6 & 60.7 & 55.0 & 1665 & 79.6 & \textbf{69.6} & 71.8 & 55.0 & 93.0\% \\
\bottomrule[1.2pt]
\end{tabular}
} \vspace{-0.35cm}
\end{table*}

\begin{table}[!t]
	\centering
	\caption{\footnotesize{Evaluation of STD on LLaVA-NeXT-7B.}}
	\label{tab:llava_next_modified}
	\vspace{0cm}
	\setlength{\tabcolsep}{0.9pt}
	\renewcommand{\arraystretch}{0.80}
	\resizebox{0.45\textwidth}{!}{
		\begin{tabular}{@{} l | *{7}{c} | c@{}}
			\toprule[1.2pt]
			\footnotesize\textbf{Method} 
			& \footnotesize\textbf{ GQA } 
			& \footnotesize\textbf{MMB} 
			& \footnotesize\textbf{MMB}$^{\text{CN}}$ 
			& \footnotesize\textbf{MME} 
			& \footnotesize\textbf{ POPE } 
			& \footnotesize\textbf{VQA}$^{\text{V2}}$ 
			& \footnotesize\textbf{VQA}$^{\text{Text}}$ 
			& \footnotesize\makecell[c]{\textbf{Avg}.} \\
			\hline

			\rowcolor{mygray}
			\multicolumn{9}{c}{\textit{Upper Bound, 2880 Tokens} \ $\textbf{(100\%)}$}   \\
			\textcolor{gray}{Vanilla} 
			& \textcolor{gray}{64.2} 
			& \textcolor{gray}{67.9} 
			& \textcolor{gray}{60.6} 
			& \textcolor{gray}{1846} 
			& \textcolor{gray}{86.4} 
			& \textcolor{gray}{81.8} 
			& \textcolor{gray}{61.3} 
			& \textcolor{gray}{100\%} \\
			\hline
			
			\rowcolor{mygray}
			\multicolumn{9}{c}{\textit{Retain 640 Tokens} \ $\fg{(\downarrow 77.8\%)}$}   \\
			PDrop & 60.6 & 65.5 & 58.5 & 1781 & 83.7 & 78.3 & \textbf{60.5} & 96.4\%  \\
			SparseVLM & 60.3 & 65.8 & 58.5 & 1773 & 84.2 & 77.1 & 59.3  & 95.9\% \\
			FiCoCo-V & 60.9 & 66.0 & 57.9 & 1793 & 84.6 & 78.7 & 58.3 & 96.2\%  \\
            V$^2$Drop & 61.1 & 65.6 & 58.4 & 1796 & 83.9 & 78.8 & 58.9 & 96.4\% \\
			\textbf{STD} & \textbf{61.8} & \textbf{66.4} & \textbf{58.9} & \textbf{1807} & \textbf{86.2} & \textbf{79.6} & 59.7 & \textbf{97.7\%}  \\
			\hline
			
			\rowcolor{mygray}
			\multicolumn{9}{c}{\textit{Retain 320 Tokens} \ $\fg{(\downarrow 88.9\%)}$}\\
			PDrop & 58.3 & 63.9 & \textbf{56.8} & 1736 & 80.2 & 75.2 & 57.8 & 93.1\%  \\
			SparseVLM & 57.7 & 63.2 & 54.4 & 1685 & 82.2 & 73.4 & 56.9  & 91.6\% \\
			FiCoCo-V & 58.8 & 63.4 & 55.3 & 1731 & 81.9 & 75.5 & 57.2 & 93.0\%  \\
            V$^2$Drop & 57.4 & 63.7 & 55.5 & 1701 & 80.2 & 74.6 & 54.6 & 91.4\% \\
			\textbf{STD} & \textbf{59.6} & \textbf{64.3} & 56.1 & \textbf{1742} & \textbf{83.2} & \textbf{76.8} & \textbf{58.3} & \textbf{94.3\%}  \\
			\hline
			
			\rowcolor{mygray}
			\multicolumn{9}{c}{\textit{Retain 160 Tokens} \ $\fg{(\downarrow 94.4\%)}$}\\
			PDrop & 54.9 & \textbf{61.8} & \textbf{54.9} & 1513 & 72.3 & 70.2 & 53.9 & 86.6\%  \\
			SparseVLM & 51.2 & 52.1 & 48.6 & 1542 & 72.7 & 66.3 & 49.2  & 80.8\% \\
			FiCoCo-V & 55.1 & 60.4 & 52.4 & 1621 & 74.5 & 70.7 & 55.3  & 87.5\% \\
            V$^2$Drop & 54.3 & 60.1 & 52.1 & 1605 & 75.2 & 70.4 & 54.6 & 86.9\% \\
			\textbf{STD} & \textbf{56.9} & 61.5 & 53.7 & \textbf{1634} & \textbf{77.3} & \textbf{72.8} & \textbf{56.6} & \textbf{89.6\%}  \\
			
			\bottomrule[1.2pt]
		\end{tabular}
	}
	\vspace{-0.10cm}
\end{table}

\subsection{Phase 3: Deep Layers via Stability-Adaptive Trigger}
\label{subsec:phase3_deep}

In deep layers ($\ell > L_2$), the model enters the \textbf{abstract semantic aggregation phase}, where attention converges to hub tokens. Our analysis shows that semantic changes in this stage are \textbf{unstable}

Instead of fixed-interval pruning, we introduce a \textit{Stability-Adaptive Trigger} to execute compression only during semantic stability. Let $\mathbf{a}^{(\ell)} \in \mathbb{R}^N$ denote the attention vector from the [CLS] token to all patch tokens at layer $\ell$. The semantic shift between consecutive layers is:
\begin{equation}
    \Delta^{(\ell)} = \left\| \mathbf{a}^{(\ell)} - \mathbf{a}^{(\ell-1)} \right\|_2.
\end{equation}

Pruning is triggered at layer $\ell$ only when the system enters a stable phase, indicated by decreasing semantic change:
\begin{equation}
    \Delta^{(\ell)} < \Delta^{(\ell-1)}.
\end{equation}
This ensures token reduction occurs only when semantic changes are diminishing, indicating that the model is entering a semantically stable state.

When pruning is triggered in this semantic aggregation phase, we use only attention weights $r_i^{(\ell)} = a_i^{(\ell)}$ to select tokens for triggered layers.

\section{Experiments}
\textbf{Experimental Settings.}
We validate STD on image understanding tasks using the LLaVA family (LLaVA-1.5-7B~\cite{liu2024improved} and LLaVA-NeXT-7B~\cite{liu2024llavanext}), Qwen2-VL~\cite{wang2024qwen2}, and the recent InternVL3-8B~\cite{zhu2025internvl3} and Qwen3-VL-8B-Instruct~\cite{bai2025qwen3vl} models. Performance is evaluated across eleven benchmarks under different token-reduction ratios. For video understanding, we conduct experiment on Video-LLaVA~\cite{lin2024video} across three benchmarks.

\textbf{Implementation Details.}
Our method is integrated into existing models in a strictly training-free manner. We use $\alpha$ to balance frequency and attention scores, setting $\alpha=0.8$ for LLaVA-1.5 and $\alpha=0.5$ for LLaVA-NeXT and Qwen2-VL. Stage-wise partitioning parameters are uniformly set to $L_1=6$ and $L_2=12$. Token pruning is performed within the vision encoder for all models and is forced to reach the target retained token number at the final pruning layer. All experiments are conducted on Nvidia RTX 4090D (24GB) GPUs.

\subsection{Effectiveness of STD}

\begin{table*}[t]
	\centering

	\begin{minipage}[t]{0.49\textwidth}
		\centering
		\caption{\footnotesize{Performance of STD on Qwen2-VL-7B-Instruct for image understanding. As the initial number of tokens varies dynamically, the reduction ratio is approximate.}}
		\label{tab:qwen2vl}
		\vspace{0cm}
		\setlength{\tabcolsep}{0.85pt}
		\renewcommand{\arraystretch}{0.80}
		\resizebox{\textwidth}{!}{
			\begin{tabular}{@{} l | *{7}{c} | c@{}}
				\toprule[1.2pt]
				\footnotesize\textbf{Method} 
				& \footnotesize\textbf{ GQA } 
				& \footnotesize\textbf{MMB} 
				& \footnotesize\textbf{MMB}$^{\text{CN}}$ 
				& \footnotesize\textbf{MME} 
				& \footnotesize\textbf{ POPE } 
				& \footnotesize\textbf{SQA} 
				& \footnotesize\textbf{VQA}$^{\text{Text}}$ 
				& \footnotesize\makecell[c]{\textbf{Avg}.} \\
				\hline

				\rowcolor{mygray}
				\multicolumn{9}{c}{\textit{Upper Bound, All Tokens} \ $\textbf{(100\%)}$}   \\
				\textcolor{gray}{Vanilla} 
				& \textcolor{gray}{61.9} 
				& \textcolor{gray}{79.8} 
				& \textcolor{gray}{79.5} 
				& \textcolor{gray}{2334} 
				& \textcolor{gray}{87.2} 
				& \textcolor{gray}{85.1} 
				& \textcolor{gray}{82.2}  
				& \textcolor{gray}{100\%} \\
				\hline
				
				\rowcolor{mygray}
				\multicolumn{9}{c}{\textit{Token Reduction} \ $\fg{(\downarrow 66.7\%)}$}\\
				FiCoCo-V & 59.4 & 78.1 & 75.8 & 2189 & 83.6 & 83.1 & 80.3  & 96.3\%  \\
				V$^2$Drop & 59.8 & 78.4 & 75.8 & 2252 & 84.0 & 83.0 & 79.8 & 96.8\% \\
				\textbf{STD} & 60.3 & 79.0 & 76.5 & 2246 & 86.1 & 83.4 & 81.0  & \textbf{97.7\%}  \\
				\hline
				
				\rowcolor{mygray}
				\multicolumn{9}{c}{\textit{Token Reduction} \ $\fg{(\downarrow 77.8\%)}$}\\
				FiCoCo-V & 57.8 & 75.6 & 73.5 & 2125 & 81.4 & 78.7 & 75.8  & 93.0\%  \\
				V$^2$Drop & 56.5 & 75.6 & 74.3 & 2139 & 81.5 & 78.9 & 73.8 & 92.4\% \\
				\textbf{STD} & 58.9 & 76.4 & 74.5 & 2187 & 84.5 & 79.0 & 76.6  & \textbf{94.5\%}  \\
				\hline
				
				\rowcolor{mygray}
				\multicolumn{9}{c}{\textit{Token Reduction} \ $\fg{(\downarrow 88.9\%)}$}\\
				FiCoCo-V & 54.7 & 70.8 & 70.1 & 2059 & 79.1 & 77.7 & 66.3 & 88.0\%  \\
				V$^2$Drop & 54.3 & 71.3 & 69.8 & 2067 & 78.6 & 77.5 & 65.3 & 87.7\% \\
				\textbf{STD} 
				& 56.3 
				& 72.1 
				& 71.8 
				& 2116 
				& 81.4 
				& 77.8 
				& 68.5  
				& \textbf{90.1\%}  \\
				
				\bottomrule[1.2pt]
			\end{tabular}
		}
	\end{minipage}
	\hfill
	\begin{minipage}[t]{0.49\textwidth}
		\centering
		\caption{\footnotesize{Performance of \textbf{STD} on video understanding tasks. Video-LLaVA uses 2048 video tokens, while our method retains 256 tokens, i.e., 32 per frame.}}
		\label{tab:video_std}
		\vspace{0cm}
		\setlength{\tabcolsep}{0.9pt}
		\renewcommand{\arraystretch}{1.25}
		\resizebox{\textwidth}{!}{%
			\begin{tabular}{@{} l |cc|cc|cc|cc @{}}
				\toprule[1.2pt]
				\multirow{2}{*}{\textbf{Method}} 
				& \multicolumn{2}{c|}{\textbf{TGIF}} 
				& \multicolumn{2}{c|}{\textbf{MSVD}} 
				& \multicolumn{2}{c|}{\textbf{MSRVT}} 
				& \multicolumn{2}{c}{\textbf{Avg.}} \\ 
				& Acc & Score & Acc & Score & Acc & Score & Acc & Score \\ 
				\hline
				
				Video-LLaVA         
				& 46.9 & 3.34 
				& 69.8 & 3.91 
				& 57.1 & 3.49   
				& 100\% & 100\% \\
				\hline

				\multirow{2}{*}{SparseVLM} 
				& 45.9 & 3.32 
				& 68.6 & 3.90 
				& 32.9 & 3.02   
				& \multirow{2}{*}{84.6\%} 
				& \multirow{2}{*}{95.2\%} \\
				& 98.9\% & 99.4\% 
				& 98.3\% & 99.7\% 
				& 57.6\% & 86.5\% 
				& & \\
				\hline

				\multirow{2}{*}{PDrop} 
				& 40.3 & 3.21 
				& 61.5 & 3.74 
				& 41.8 & 3.19  
				& \multirow{2}{*}{82.4\%} 
				& \multirow{2}{*}{94.4\%} \\
				& 85.9\% & 96.1\% 
				& 88.1\% & 95.7\% 
				& 73.2\% & 91.4\% 
				& & \\
				\hline

				\multirow{2}{*}{FiCoCo-V} 
				& 43.4 & 3.25 
				& 67.9 & 3.89 
				& 48.3 & 3.26  
				& \multirow{2}{*}{91.4\%} 
				& \multirow{2}{*}{96.7\%} \\
				& 92.5\% & 97.3\% 
				& 97.2\% & 99.5\% 
				& 84.6\% & 93.4\% 
				& & \\
				\hline

				\multirow{2}{*}{\textbf{STD}} 
				& 45.7 & 3.32 
				& 67.6 & 3.88 
				& 54.9 & 3.44 
				& \multirow{2}{*}{\textbf{96.7\%}}
				& \multirow{2}{*}{\textbf{99.1\%}} \\
				& 97.4\% & 99.4\% 
				& 96.8\% & 99.2\% 
				& 96.1\% & 98.6\% 
				& & \\
				
				\bottomrule[1.2pt]
			\end{tabular}%
		}
	\end{minipage}

	\vspace{-0.1cm}
\end{table*}

\begin{table*}[t]
	\centering
	\caption{\textbf{Performance of STD on LLaVA-1.5-7B (left) and LLaVA-NeXT-7B (right).} \(\Delta\) denotes the inference speedup factor. All experiments used the POPE benchmark on a single NVIDIA 4090D GPU.}
    \vspace{0cm}
	\label{tab:efficiency}
	\begin{minipage}[t]{0.49\linewidth}
		\centering
		\setlength{\tabcolsep}{2.5pt}
		\renewcommand{\arraystretch}{0.92}
		\footnotesize
		\resizebox{\linewidth}{!}{
			\begin{tabular}{lcccccc}
				\toprule
				\multirow{2}{*}{\textbf{Methods}} & 
				\multirow{2}{*}{\textbf{Token}} & 
				\textbf{Total} & 
				\multirow{2}{*}{\textbf{\(\Delta\)$\uparrow$}} & 
				\textbf{Prefilling}  & 
				\multirow{2}{*}{\textbf{\(\Delta\)$\uparrow$}} &
				\multirow{2}{*}{\textbf{TFLOPs}}\\
				& & 
				\textbf{\hspace{5pt}Time$\downarrow$} &  & 
				\textbf{ Time$\downarrow$} &  &\\
				\midrule
				\textcolor{gray}{LLaVA-1.5-7B} & \textcolor{gray}{576} & \textcolor{gray}{17:53} & \textcolor{gray}{1.0$\times$}& \textcolor{gray}{118.0ms} & \textcolor{gray}{1.0$\times$}&\textcolor{gray}{10.20}\\
				+ SparseVLM & 64 & 16:15 & 1.1$\times$ & 107.3ms & 1.1$\times$ &2.67 \\
				+ PDrop & 64 & 13:45 & 1.3$\times$ & 84.3ms & 1.4$\times$ &2.49 \\
                + FlowCut& 64 & 11:55 & 1.5$\times$ & 73.8ms & 1.6$\times$  &2.25 \\
                + V$^2$Drop& 64 & 12:51 & 1.4$\times$ & 79.2ms & 1.5$\times$  &2.39 \\
				\rowcolor{lightgreen!80} 
				\textbf{+ STD}& 64 & \textbf{10:05} & \textbf{1.7$\times$} & \textbf{64.3ms} & \textbf{1.8$\times$}  &\textbf{1.99} \\
				\bottomrule
			\end{tabular}
		}
	\end{minipage}
	\hfill 
	\begin{minipage}[t]{0.49\linewidth}
		\centering
		\setlength{\tabcolsep}{1.85pt}
		\renewcommand{\arraystretch}{0.92}
		\footnotesize
		\resizebox{\linewidth}{!}{
			\begin{tabular}{lcccccc}
				\toprule
				\multirow{2}{*}{\textbf{Methods}} & 
				\multirow{2}{*}{\textbf{Token}} & 
				\textbf{Total} & 
				\multirow{2}{*}{\textbf{\(\Delta\)$\uparrow$}} & 
				\textbf{Prefilling}  & 
				\multirow{2}{*}{\textbf{\(\Delta\)$\uparrow$}} &
				\multirow{2}{*}{\textbf{TFLOPs}}\\
				& & 
				\textbf{\hspace{5pt}Time$\downarrow$} &  & 
				\textbf{ Time$\downarrow$} &  &\\
				\midrule
				\textcolor{gray}{LLaVA-NeXT-7B} & \textcolor{gray}{2880} & \textcolor{gray}{50:24} &\hspace{2pt}\textcolor{gray}{1.0$\times$}& \textcolor{gray}{323.0ms} & \textcolor{gray}{1.0$\times$}&\textcolor{gray}{45.61}\\
				+ SparseVLM & 160 & {42:00} &\hspace{2pt}1.2$\times$ & 215.3ms & 1.5$\times$ & 11.20 \\
				+ PDrop & 160 & {19:23} &\hspace{2pt}2.6$\times$ & 124.2ms & 2.6$\times$ &9.96 \\
                + FlowCut& 160 & 16:48 &\hspace{2pt}3.0$\times$ & 100.9ms & 3.2$\times$  &6.30\\
                + V$^2$Drop& 160 & 18:10 &\hspace{2pt}2.8$\times$ & 114.7ms & 2.8$\times$  &8.97\\
				\rowcolor{lightgreen!80} 
				\textbf{+ STD}& 160 & \textbf{13:07} &\hspace{2pt}\textbf{3.8$\times$} & \textbf{83.2ms} & \textbf{3.9$\times$}  &\textbf{5.16}\\
				\bottomrule[1.1pt]
			\end{tabular}
		}
	\end{minipage}
    \vspace{-0.10cm}
\end{table*}

\paragraph{\textbf{Image understanding tasks}} We evaluate STD on LLaVA-1.5-7B in a training-free inference setting. As shown in Table~\ref{tab:main}, STD outperforms existing methods by 1.1\% with 64 tokens, and by 1.8\% when combined with Flowcut. With 192 retained tokens, STD + SparseVLM surpasses the SOTA by 1.5\% and the original SparseVLM by 3.6\%. On LLaVA-NeXT-7B, Table~\ref{tab:llava_next_modified} shows that STD exceeds the state-of-the-art by 2.1\% with 160 tokens. We also evaluate Qwen2-VL, and Table~\ref{tab:qwen2vl} shows that STD consistently performs best across different token reduction ratios on Qwen2-VL-7B.

\vspace{-0.2cm}
\paragraph{Video understanding tasks}
To evaluate \textbf{STD} for video understanding, we integrate it into Video-LLaVA. Each video has 8 frames with 256 tokens per frame (2048 total), compressed to 256 tokens, i.e., 32 per frame. As shown in Table~\ref{tab:video_std}, \textbf{STD} remains robust, retaining 96.7\% average accuracy on TGIF, MSVD, and MSRVT. It outperforms the previous state-of-the-art FiCoCo-V by 5.3\%, showing its ability to preserve key information while reducing computation for multimodal video reasoning.

\vspace{-0.2cm}
\paragraph{\textbf{Efficiency Analysis}} Our proposed STD improves the inference efficiency of LVLMs . Table~\ref{tab:efficiency} compares total inference time, prefilling time, and FLOPs of our method with existing approaches on LLaVA-1.5-7B and LLaVA-NeXT-7B. STD consistently achieves the best efficiency gains, including a \textbf{3.9× speedup} in prefilling and a \textbf{3.8× speedup} in overall inference on LLaVA-NeXT.

\vspace{-0.2cm}

\begin{figure*}[t]
\centering
\includegraphics[width=0.92\linewidth]{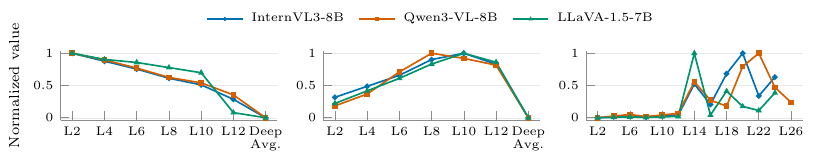}
\vspace{-0.20cm}
\captionof{figure}{Generalization of stage-wise findings on recent MLLMs. From left to right: edge similarity, object similarity, and adjacent-layer attention shift. Each series is min--max normalized within its model and metric. ``Deep Avg.'' averages L14 through the final layer in the first two plots. Recent MLLMs exhibit the same pattern with LLaVA-1.5: edge similarity peaks in shallow layers, object similarity peaks in intermediate layers, and semantic representations shift markedly in deep layers.}
\label{fig:recent_stage_analysis}
\vspace{-0.08cm}
\end{figure*}

\subsection{Ablation Studies}

\newcommand{\ablationtables}{%
\begin{table*}[t]
    \centering 
    
    \begin{minipage}[t]{0.58\textwidth} 
        \caption{Ablation study on component effectiveness.}
        \label{tab:ablation}
        \vspace{-0.08cm}
        \setlength{\tabcolsep}{2.5pt}
        \renewcommand{\arraystretch}{1.05}
        \resizebox{\linewidth}{!}{
            \begin{tabular}{l|ccccccc}
                \toprule
                \small\textbf{Methods} & 
                \small\textbf{Token} & 
                \small\textbf{HSA}  & 
                \small\textbf{GSA} & 
                \small\textbf{SAT}  & 
                \small\textbf{TextVQA} & 
                \small\textbf{GQA} &
                \small\textbf{ POPE } 
                \\
                \midrule
                \textcolor{gray}{LLaVA-1.5-7B} & \textcolor{gray}{576} & \textcolor{gray}{--}& \textcolor{gray}{--} & \textcolor{gray}{--}& \textcolor{gray}{58.3} & \textcolor{gray}{61.9}&\textcolor{gray}{85.9}\\
                \hline
                Standard Multi-Layer Pruning  & 64 & - & - & - &54.5 &54.7  &79.2\\
                Standard Multi-Layer Pruning(w/o Shallow) & 64 & - & - & - &55.0 &55.1  &79.8\\
                \hline
                    STD(a)  & 64 &\checkmark&  & &55.5 &55.6  &80.3\\
                STD(b) & 64 &&\checkmark  &  &55.2 &55.4 &79.9\\
                STD(c)  & 64 &&  & \checkmark &55.4 &55.6&80.6\\
                \hline
                \rowcolor{lightgreen!80}
                \textbf{STD} & 64 &\checkmark & \checkmark & \checkmark & \textbf{56.0} & \textbf{56.1}  &\textbf{81.2} \\
                \bottomrule
            \end{tabular}
        }
    \end{minipage}
    \hfill 
    \begin{minipage}[t]{0.40\textwidth} 
        \caption{Performance comparison of deploying components at different layer depths.}
        \label{tab:density_layers}
        \vspace{-0.08cm}
        \setlength{\tabcolsep}{3pt}
        \renewcommand{\arraystretch}{0.65}
        \resizebox{\linewidth}{!}{
            \begin{tabular}{c|ccc|cc}
                \toprule
                \textbf{Component} & \textbf{Shallow} & \textbf{Intermediate} & \textbf{Deep} & \textbf{TextVQA} & \textbf{POPE} \\
                \midrule
                \multirow{3}{*}{HSA}
                    & $\checkmark$ & & &\textbf{55.5} & \textbf{80.3}\\
                    & & $\checkmark$ & & 54.1& 79.0\\
                    & & & $\checkmark$ & 51.6& 72.7\\
                \hline
                \multirow{3}{*}{GSA}
                    & $\checkmark$ & & & 53.9& 78.6\\
                    & & $\checkmark$ & & \textbf{55.2}& \textbf{79.9}\\
                    & & & $\checkmark$ &53.2 & 77.8\\
                \hline
                \multirow{3}{*}{SAT}
                    & $\checkmark$ & & & 54.3& 78.8\\
                    & & $\checkmark$ & & 54.7& 79.6\\
                    & & & $\checkmark$ & \textbf{55.4}& \textbf{80.6}\\
                \bottomrule
            \end{tabular}
        }
    \end{minipage}
    \vspace{-0.05cm}
\end{table*}
}

\textbf{Effectiveness of Each Component.}
Table~\ref{tab:ablation} reports the ablation results on three benchmarks with 64 retained tokens. Adding any single component, \textit{High-Frequency Spectral Analysis}(HSA), \textit{Gaussian-Smoothed Attention}(GSA), or \textit{Stability-Adaptive Trigger}(SAT), consistently improves over the baseline variants. Using all three yields the best results, showing that each module is effective and most effective when combined.

\textbf{Importance of Layer-Specific Deployment.}
Our design uses \textit{HSA} in shallow layers, \textit{GSA} in intermediate layers, and \textit{SAT} in deep layers. Table~\ref{tab:density_layers} shows that changing this assignment degrades performance. For example, applying HSA module to deeper layers lowers the POPE score by up to 7.6\% than our shallow-layer setup. These results show the importance of layer-specific strategies.

\ablationtables

\vspace{-0.25cm}

\subsection{Generalization to Recent MLLMs}
\label{sec:recent_mllms}

We examine whether our stage-wise findings generalize to recent visual encoders and STD remains effective on these models. We use token-level attention as a proxy for visual focus. For the shallow-layer analysis, Canny responses are aggregated into token-level edge references. For the intermediate-layer analysis, the main subjects of 1,024 images are manually annotated and converted into token-level subject references. We report cosine similarity between these references and layer-wise attention. For deep layers, we report the $L_2$ distance between adjacent attention maps ($\times 0.01$). For visualization, each model's curve is independently min--max normalized within each metric.

\newcommand{\analysisfigure}{%
\begin{figure*}[!t]
    \centering
    \includegraphics[width=0.76\linewidth]{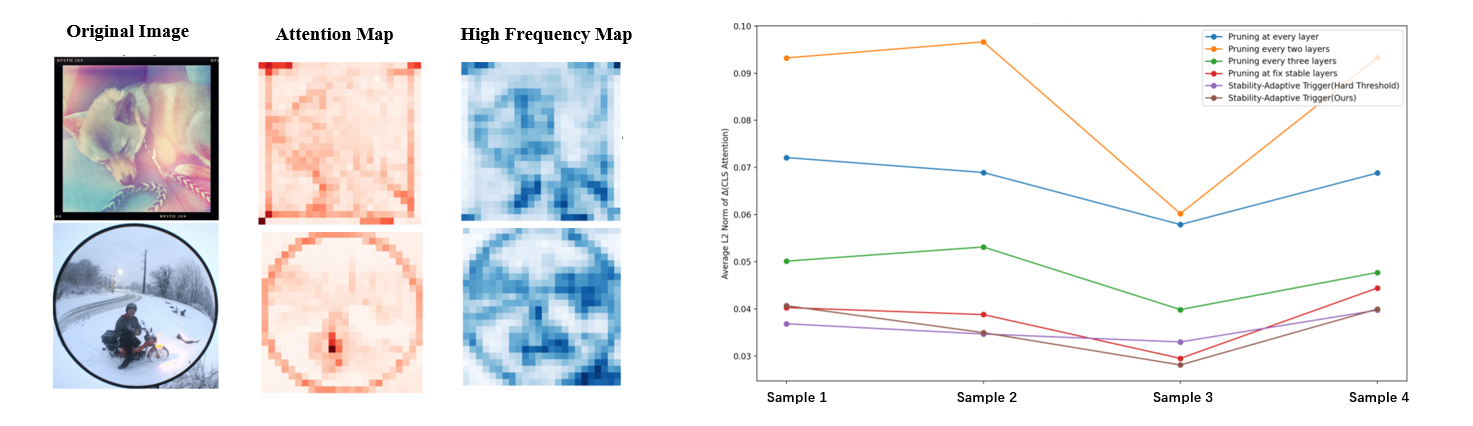}
    \vspace{-0.05cm}
    \caption{
\textbf{(Left)} Shallow-layer attention versus High-Frequency Spectral Analysis; the spectral representation provides a deterministic proxy for structural contours. \textbf{(Right)} Deep-layer pruning strategies; our Stability-Adaptive Trigger adaptively selects more stable layers than uniform-interval pruning.}
    \label{fig:ana_diss}
    \vspace{-0.25cm}
\end{figure*}
}

\newcommand{\recentmodelstable}{%
\begin{table*}[!t]
\centering
\captionof{table}{Comparison with state-of-the-art methods on recent MLLMs at a 90.0\% token reduction rate. Our method consistently achieves the best performance.}
\label{tab:recent_mllm_eval}
\vspace{-0.08cm}
\begin{minipage}[t]{0.49\textwidth}
\centering
\setlength{\tabcolsep}{3.4pt}
\renewcommand{\arraystretch}{0.62}
\resizebox{0.84\linewidth}{!}{
\begin{tabular}{lrrrrr}
\toprule
\multicolumn{6}{c}{\textbf{InternVL3-8B}}\\
\textbf{Method} & \textbf{MME} & \textbf{MMB} & \textbf{POPE} & \textbf{SQA} & \textbf{Avg.}\\
\midrule
Vanilla & 2369 & 85.7 & 90.4 & 97.9 & 100\%\\
FlowCut & 1978 & 77.9 & 86.3 & 85.8 & 89.4\%\\
FiCoCo-V & 1869 & 74.8 & 82.7 & 81.5 & 85.2\%\\
V$^2$Drop & 1994 & 78.1 & 83.4 & 86.3 & 88.9\%\\
\textbf{STD} & \textbf{2034} & \textbf{78.3} & \textbf{87.5} & \textbf{86.8} & \textbf{90.7\%}\\
\bottomrule
\end{tabular}}
\end{minipage}
\hfill
\begin{minipage}[t]{0.49\textwidth}
\centering
\setlength{\tabcolsep}{3.4pt}
\renewcommand{\arraystretch}{0.68}
\resizebox{\linewidth}{!}{
\begin{tabular}{lrrrrr}
\toprule
\multicolumn{6}{c}{\textbf{Qwen3-VL-8B-Instruct}}\\
\textbf{Method} & \textbf{MIA-Bench} & \textbf{MMB} & \textbf{MMStar} & \textbf{RWQA} & \textbf{Avg.}\\
\midrule
Vanilla & 91.1 & 85.0 & 70.9 & 71.5 & 100\%\\
FlowCut & 87.2 & 77.5 & 54.4 & 60.6 & 87.1\%\\
FiCoCo-V & 83.9 & 76.1 & 50.2 & 56.6 & 82.9\%\\
V$^2$Drop & 84.6 & 77.0 & 52.4 & 57.8 & 84.5\%\\
\textbf{STD} & \textbf{88.0} & \textbf{77.2} & \textbf{56.7} & \textbf{62.7} & \textbf{88.8\%}\\
\bottomrule
\end{tabular}}
\end{minipage}
\vspace{-0.20cm}
\end{table*}
}

Figure~\ref{fig:recent_stage_analysis} exhibits the same qualitative pattern across both recent architectures and the LLaVA-1.5 reference: edge similarity is highest in shallow layers, object similarity peaks at intermediate depths, and the variation between adjacent attention maps remains consistently low in shallow and intermediate layers but fluctuate sharply in deep layers. These results demonstrate that our stage-wise findings also hold for recent models. To further evaluate its generalization, we apply STD to the recent models and report the results in Table~\ref{tab:recent_mllm_eval}. STD achieves the best average performance among the compared pruning methods on both models, outperforming the SOTA method by $1.3\%$ and $1.7\%$, respectively, demonstrating that our method remains highly effective on recent models.

\subsection{Complementarity with Projector/LLM-Stage Pruning}
\label{sec:complementarity}

Visual token pruning can be performed at different stages of the multimodal pipeline, including the vision encoder, projector, and LLM. Vision-encoder pruning identifies redundancy from a visual-perception perspective, whereas projector/LLM-stage pruning targets semantic redundancy and can leverage textual guidance for token selection. We find that vision-encoder pruning and projector/LLM-stage pruning are complementary. To verify this fairly, we apply STD within the vision encoder and subsequently apply DART~\cite{wen2025dart} or V$^2$Drop~\cite{chen2025v2drop} at downstream stages.

\newcommand{\complementaritytable}{%
\begin{figure}[t]
\centering
\captionof{table}{Complementarity of pruning stages on LLaVA-1.5-7B. Combining STD(Vision Encoder Stage) with Projector/LLM Stage pruning achieves better performance than either approach alone.}
\label{tab:stage_complementarity}
\setlength{\tabcolsep}{2.0pt}
\resizebox{0.90\columnwidth}{!}{
\begin{tabular}{llrrrrr}
\toprule
\textbf{Method} & \textbf{Stage} & \textbf{GQA} & \textbf{POPE} & \textbf{SQA} & \textbf{TextVQA} & \textbf{Avg.}\\
\midrule
Original & None & 61.9 & 85.9 & 69.5 & 58.3 & 100\%\\
STD & Vision & 58.8 & 84.6 & 68.8 & 57.4 & 97.7\%\\
DART & Projector & 57.9 & 80.1 & 69.1 & 56.4 & 95.7\%\\
STD+DART & Vision+Projector & 59.3 & 85.0 & 69.1 & 57.8 & \textbf{98.3\%}\\
V$^2$Drop & LLM & 56.3 & 80.9 & 68.8 & 53.8 & 94.1\%\\
STD+V$^2$Drop & Vision+LLM & \textbf{59.5} & 84.8 & \textbf{69.3} & 57.6 & \textbf{98.3\%}\\
\bottomrule
\end{tabular}}
\vspace{-0.05cm}
\end{figure}
}

Both combinations substantially outperform the corresponding individual methods and achieve the best average scores (Table~\ref{tab:stage_complementarity}). These results show that pruning at different stages addresses distinct sources of redundancy and can be effectively combined. The SparseVLM+STD results in Table~\ref{tab:main} further support this observation.

\vspace{-0.05cm}

\subsection{Analysis and Discussion}
\vspace{-0.05cm}

\recentmodelstable
\analysisfigure
\complementaritytable

\textbf{Effectiveness of High-Frequency Spectral Analysis in Shallow Layers.} As shown in \textbf{Fig.~\ref{fig:ana_diss}}(Left), we visualize shallow-layer attention maps and our High-Frequency Spectral Analysis results. Standard shallow-layer attention can detect edges, but is often uncertain and chaotic. In contrast, our spectral analysis offers a more deterministic proxy to detect edge contours. This is further supported by the quantitative results in \textbf{Fig.~\ref{fig:combined_attn_analysis}}(Middle).

\textbf{Validation of Stability-Adaptive Triggering Strategy.}
To validate the Stability-Adaptive Trigger, we plot its selected deep pruning layers under different strategies in \textbf{Fig.~\ref{fig:ana_diss}}(Right). For each strategy, we compute the selected layers' average stability metric to evaluate pruning stability. We compare uniform intervals (every $k$ layers) with stability-based approaches. The results show that uniform pruning often selects layers with high semantic fluctuation, causing instability. Among the stability-aware strategies, the selected pruning layers are stable and yield similar average stability metrics. However, our approach avoids re-selecting fixed layers or re-tuning thresholds for different model architectures, providing better adaptability.

\section{Related Work}
\enlargethispage{\baselineskip}
\vspace{-0.1cm}
\label{sec:related}
\noindent\textbf{Projector/LLM Stage Pruning.}
Token pruning can be performed in the Projector or LLM stage. Its main advantage is that pruning can be guided by cross-modal attention in LLM layers (e.g., SparseVLM~\cite{zhang2024sparsevlm}). Many studies~\cite{xing2024pyramiddrop,zhang2025adaptinfer,chen2025v2drop,ye2025atp,li2025transprune} progressively remove redundant tokens from shallow to deep LLM layers. For example, PDrop~\cite{xing2024pyramiddrop} divides LLM layers into stages and drops tokens with low text-vision similarity at predefined ratios in selected layers. Some advanced methods~\cite{hu2024illava,zhang2025vscan,wu2026hidrop} further analyze modality interaction patterns across layers and design stage-specific strategies.
However, these methods have several distinct drawbacks: (1) the Vision Encoder still incurs high computational costs before pruning; (2) for LLM-stage pruning, sending all visual tokens into early LLM layers causes heavy VRAM usage; and (3) they are unfriendly to multi-round conversations, since visual tokens must be recompressed for each new query.

\noindent\textbf{Vision Encoder Pruning.}
Pruning within the Vision Encoder offers larger gains in both speed and memory efficiency, as redundant tokens are removed before entering the LLM. A notable branch of work~\cite{tong2025flowcut,han2026filter,chen2026evoprune} progressively removes redundancy from shallow to deep Vision Encoder layers. These methods also recognize the limitations of attention-only criteria. For example, FlowCut~\cite{tong2025flowcut} combines \textit{Semantic Similarity} and \textit{Information Density} with attention, while FiCoCo-V~\cite{han2026filter} further considers a \textit{Local Penalty Strategy} and \textit{Correlation-Based Information Recycling}. Although these multi-metric methods partly reduce the errors caused by pure attention-based pruning, they still apply the same criteria across all layers without considering the distinct functional characteristics and attention patterns of shallow and deep Vision Encoder layers.

\section{Conclusion}
\label{sec:conclusion}
In this work, we present empirical evidence for a recurring stage-wise pattern: shallow representations emphasize edges, intermediate representations align with object-level regions, and deep representations exhibit non-uniform attention shifts consistent with semantic aggregation. Motivated by this hypothesis, we propose STD, a hierarchical pruning framework that preserves edges in shallow layers, enhances spatial coherence in intermediate layers, and avoids pruning during unstable deep-layer transitions. Extensive experiments across model families show the effectiveness of STD.

\clearpage
\section*{Acknowledgments}
This work is supported by the National Natural Science Foundation of China under grants 62206102; the National Key Research and Development Program of China under grant 2024YFC3307900; the National Natural Science Foundation of China under grants 62436003, 62376103 and 62302184; Major Science and Technology Project of Hubei Province under grant 2025BAB011 and 2024BAA008; Hubei Science and Technology Talent Service Project under grant 2024DJC078; Ant Group through CCF-Ant Research Fund; 
\section*{Limitations}
Although STD achieves consistent improvements across different LVLMs and tasks, it still has several limitations. First, our method is built on an empirical characterization of stage-wise functional roles in the vision encoder, and its effectiveness may depend on the architectural properties of the underlying visual backbone. Whether the same shallow-to-deep transition pattern holds universally for other encoders or multimodal architectures requires further investigation. Second, while STD is training-free and plug-and-play, it still introduces additional hand-crafted components, such as spectral analysis, Gaussian smoothing, and stability-based triggering, whose hyperparameters may require adjustment under different resolutions, token budgets, or model scales. Third, although STD significantly reduces prefilling cost, it is designed mainly for visual token redundancy in inference and does not directly address other efficiency bottlenecks, such as KV-cache growth or decoding latency in the LLM stage. Finally, our current study mainly validates the method on image and video understanding benchmarks; its robustness on more diverse multimodal settings, such as document understanding, embodied perception, or highly fine-grained visual reasoning, remains an important direction for future work.

\bibliography{custom}
\clearpage
\appendix

\section{Dataset}\label{dataset}
Our approach was rigorously evaluated using thirteen distinct benchmarks: ten dedicated to image understanding and three focused on video understanding. Each benchmark targets specific dimensions of multimodal intelligence.

\textbf{GQA}~\cite{hudson2019gqa}
Structured around scene graphs, queries, and corresponding images, the GQA benchmark enriches visual data with detailed spatial relationships and object attributes. Its questions are specifically crafted to test a model's capacity for scene comprehension and reasoning across various image aspects.

\textbf{MMBench}~\cite{liu2024mmbench}
MMBench assesses model performance through a three-tiered hierarchy. The initial level (L-1) examines basic perception and reasoning skills. The second tier (L-2) broadens this scope into six sub-abilities, while the final level (L-3) further delineates these into 20 precise dimensions. This layered architecture facilitates a granular and thorough evaluation of diverse model capabilities. The benchmark also includes MMB-CN, its Chinese-language counterpart.

\textbf{MME}~\cite{fu2023mme}
The MME benchmark provides a comprehensive assessment of perceptual and cognitive faculties across 14 subtasks. By utilizing manually curated instruction-answer pairs alongside concise prompts, it effectively mitigates data leakage risks, ensuring a fair and accurate measurement of model performance.

\textbf{POPE}~\cite{li2023evaluating}
POPE is designed to systematically detect object hallucinations via binary queries regarding object presence in images. Employing metrics such as accuracy, recall, precision, and F1 score, it offers a precise quantification of hallucination rates under various sampling strategies.

\textbf{ScienceQA}~\cite{lu2022learn}
Covering a broad spectrum of domains including natural, language, and social sciences, ScienceQA organizes its queries hierarchically into 26 topics, 127 categories, and 379 distinct skills. This extensive structure provides a diverse set of scientific problems, effectively testing multimodal comprehension, multi-step reasoning, and model interpretability.

\textbf{VQA-V2}~\cite{goyal2017making}
VQA-V2 tests visual perception through open-ended inquiries based on 265,016 real-world images. With each question accompanied by ten human-annotated ground truth answers, the benchmark allows for a robust evaluation of a model's ability to interpret and respond to visual questions.

\textbf{TextVQA}~\cite{singh2019towards}
TextVQA centers on the synergy between visual elements and embedded text within images. It challenges models to simultaneously process visual cues and textual content to accurately answer questions, thereby evaluating integrated visual-textual understanding.

\textbf{MIA-Bench}~\cite{qian2025mia}
MIA-Bench evaluates whether multimodal large language models can strictly follow complex, layered instructions grounded in images. It contains 400 image--prompt pairs with manually written instructions comprising multiple sub-instructions, enabling fine-grained assessment of both visual understanding and instruction adherence.

\textbf{MMStar}~\cite{chen2024mmstar}
MMStar is a vision-indispensable multimodal benchmark containing 1,500 human-curated samples. Its evaluation spans six core capabilities and 18 detailed axes, with samples selected to require visual input and to reduce the effects of data leakage and text-only shortcuts.

\textbf{RWQA}~\cite{xai2024realworldqa}
RealWorldQA (RWQA) evaluates understanding of real-world visual scenes. Its initial release contains more than 700 images, including anonymized images captured from vehicles and other real-world images, with each image paired with a question and an easily verifiable answer.

\textbf{TGIF-QA}~\cite{jang2017tgif}
Extending question answering to the video domain, TGIF-QA utilizes 165,000 QA pairs derived from GIFs. It defines four task categories: three that demand spatio-temporal reasoning (counting repetitions, identifying repeating actions, and tracking state transitions) and one frame-based QA task solvable from single images.

\textbf{MSVD-QA}~\cite{xu2017video}
Built upon the MSVD dataset, the MSVD-QA benchmark includes 1,970 video clips paired with approximately 50.5K question-answer sets. It features open-ended questions spanning five categories (what, who, how, when, where), covering varied aspects of video content to support both video QA and captioning evaluations.

\textbf{MSRVTT-QA}~\cite{xu2017video}
Comprising 10,000 video clips and 243,000 QA pairs, MSRVTT-QA challenges models to synthesize visual and temporal information. Mirroring the structure of MSVD-QA, it incorporates five question types to assess a model's proficiency in understanding complex dynamic video content.

\section{Evaluation under less token budgets and Comparison with More Methods} \label{More_method}

\paragraph{Retain less visual tokens} 
To further validate the robustness of our method under highly constrained token budgets, we conduct experiments retaining only 32 visual tokens. As shown in Table~\ref{tab:less_token}, our method achieves superior performance compared to existing approaches under this aggressive compression setting. Notably, the performance gain of STD over FlowCut in this 32-token setting is even more pronounced than in the 64-token scenario, demonstrating that our approach is particularly effective at preserving critical informative content when the token budget is severely limited.

\begin{table}[!t]
	\vspace{-0.15cm}
	\centering
	\caption{Performance of STD on LLaVA-1.5 under 32 visual token setting.}
	\label{tab:less_token}
	\setstretch{1.05} 
	\setlength{\tabcolsep}{4pt}
	\vspace{-0.05cm}
	\resizebox{0.9\linewidth}{!}{
		\begin{tabular}{l | c | c c c c c c c c | c}
			\toprule[1.2pt]
			\small\textbf{Method} & \small\textbf{Tokens} & \small\textbf{GQA} & \small\textbf{MMB} & \small\textbf{MMB-CN} & \small\textbf{MME} & \small\textbf{POPE} & \small\textbf{SQA} & \small\textbf{VQAv2} & \small\textbf{TextVQA} & \small\textbf{Avg} \\
			\hline
			LLaVA-1.5 7B (upper limit) & 576 & 61.9 & 64.7 & 58.1 & 1865 & 85.9 & 69.5 & 78.5 & 58.3 & 100\% \\
			\hline
            FlowCut & 32 & 52.1 & 57.0 & 50.3 & 1612 & 69.6 & 68.7 & 66.9 & 53.0 & 87.6\% \\
            FiCoCo-V & 32 & 51.8 & 55.9 & 50.0 & 1534 & 69.4 & 68.4 & 65.7 & 52.3 & 86.3\% \\
			STD (Ours) & 32 & 53.8 & 58.4 & 51.5 & 1599 & 74.8 & 68.6 & 69.3 & 54.3 & \textbf{89.8\%} \\
			\bottomrule[1.2pt]
		\end{tabular}
	}
    \vspace{-0.4cm}
\end{table}

\paragraph{Compare with more methods}
In addition to the representative methods discussed in the main text, we extend our comparison to include several recent works listed in Table~\ref{tab:more_method_updated}: 
FastV~\cite{chen2024image} is one of the earliest explorations into vision token pruning for multimodal models, identifying significant token redundancy in the deep layers of the vision encoder and proposing a method that leverages deep-layer attention mechanisms to perform effective token pruning. 
DART~\cite{wen2025dart} challenges traditional importance-based pruning by instead selecting tokens with low duplication relative to a small set of pivot tokens. 
DivPrune~\cite{alvar2025divprune} formulates token selection as a Max-Min Diversity Problem to maximize the diversity of retained visual tokens, thereby minimizing redundancy. 
TRIM~\cite{song2025trim} mimics human attention patterns in Visual Question Answering by utilizing CLIP-based metrics to identify and retain only the most semantically relevant image tokens.

Most existing methods either apply importance metrics (e.g., attention in FastV, diversity in DivPrune, or semantic relevance in TRIM) at a single layer or employ uniform criteria across multiple layers without differentiation, thereby overlooking the distinct evolutionary characteristics of the model from shallow to deep stages. In contrast, our STD framework pioneers a \textit{stage-aware} paradigm that adapts to the intrinsic evolution of visual features: utilizing High-Frequency Spectral Analysis for noisy shallow layers, Gaussian-Smoothed Attention for coherent intermediate regions, and a Stability-Adaptive Trigger for deep semantic aggregation. As shown in Table~\ref{tab:more_method_updated}, this adaptive approach yields superior comprehensive performance, achieving the highest average score (\textbf{93.5}) among all compared methods.
\vspace{-0.15cm}
\begin{table}[!t]
	\centering
	\caption{Comparison of methods on LLaVA-1.5-7B with various benchmarks.}
	\label{tab:more_method_updated}
	\vspace{-0.05cm}
	\setlength{\tabcolsep}{4pt}
	\setstretch{1.05}
	\resizebox{0.98\linewidth}{!}{
		\begin{tabular}{l | c c c c c c | c}
			\toprule[1.2pt]
			\textbf{Method} & \textbf{GQA} & \textbf{MMB} & \textbf{MMBCN} & \textbf{POPE} & \textbf{VQAv2} & \textbf{TextVQA} & \textbf{Avg} \\
			\hline
			LLaVA-1.5 7B (Upper Limit) & 61.9 & 64.7 & 58.1 & 85.9 & 78.5 & 58.2 & 100\% \\
			\hline
			FastV~(ECCV24) & 46.1 & 48.0 & 52.7 & 48.0 & 55.0 & 47.8 & 74.6\% \\
			DART~(EMNLP25) & 54.7 & 59.5 & 54.0 & 73.8 & 71.3 & 54.7 & 90.7\% \\
			DivPrune~(CVPR25) & 57.2 & 60.1 & 51.9 & 85.2 & 74.0 & 54.0 & 93.5\% \\
			TRIM~(COLING25) & 56.9 & 61.5 & 44.9 & 86.7 & 71.6 & 50.0 & 90.4\% \\
			\textbf{STD (Ours)} & 56.1 & 61.1 & 55.3 & 81.6 & 73.3 & 56.0 & \textbf{94.1\%} \\
			\bottomrule[1.2pt]
		\end{tabular}
	}
\end{table}

\section{Sensitivity Analyses of Hyper Parameters}

To report the STD parameters and examine their robustness, we conduct a sensitivity analysis on LLaVA-NeXT-7B and Qwen2.5-VL-7B using POPE. We study three key parameters: $\alpha$, which balances frequency and attention scores; $(L_1,L_2)$, which defines the stage-wise partition boundaries; and $\rho$, which specifies the proportion of high-frequency bands. When varying one parameter, we fix the others at $\alpha=0.5$, $(L_1,L_2)=(6,12)$, and $\rho=1/3$.

As shown in Table~\ref{tab:parameter_sensitivity}, both models are robust to these choices. Relative to the default $\alpha=0.5$, setting $\alpha=1.0$ changes the POPE score by only 0.2 for LLaVA-NeXT-7B and 0.3 for Qwen2.5-VL-7B. The sweeps over $(L_1,L_2)$ and $\rho$ similarly yield small variations, demonstrating that STD is not sensitive to the precise parameter settings within the tested ranges.

\begin{table*}[!t]
	\centering
	\caption{Sensitivity analysis of STD on POPE. When one parameter is varied, the others are fixed at $\alpha=0.5$, $(L_1,L_2)=(6,12)$, and $\rho=1/3$. Bold indicates the best score within each model and parameter sweep.}
	\label{tab:parameter_sensitivity}
	\setstretch{1.05}
	\setlength{\tabcolsep}{7pt}
	\resizebox{0.96\textwidth}{!}{
		\begin{tabular}{llccccccc}
			\toprule[1.2pt]
			\textbf{Parameter} & \textbf{Model} & \multicolumn{7}{c}{\textbf{Parameter values and POPE scores}} \\
			\midrule
			$\alpha$ & & 0 & 0.2 & 0.4 & 0.5 & 0.6 & 0.8 & 1.0 \\
			& LLaVA-NeXT-7B & 82.1 & 82.4 & 82.9 & \textbf{83.2} & \textbf{83.2} & 83.1 & 83.0 \\
			& Qwen2.5-VL-7B & 83.5 & 83.8 & 84.2 & \textbf{84.5} & 84.4 & 84.2 & 84.2 \\
			\midrule
			$(L_1,L_2)$ & & $(2,8)$ & $(4,10)$ & $(4,12)$ & $(6,12)$ & $(6,14)$ & $(8,14)$ & $(8,16)$ \\
			& LLaVA-NeXT-7B & 82.3 & 82.6 & 83.0 & \textbf{83.2} & 83.1 & 82.7 & 82.4 \\
			& Qwen2.5-VL-7B & 83.7 & 84.1 & 84.3 & \textbf{84.5} & \textbf{84.5} & 83.9 & 83.5 \\
			\midrule
			$\rho$ & & $1/6$ & $1/4$ & $1/3$ & $1/2$ & $2/3$ & $3/4$ & $5/6$ \\
			& LLaVA-NeXT-7B & 82.7 & 83.1 & \textbf{83.2} & 83.0 & 83.0 & 82.8 & 82.5 \\
			& Qwen2.5-VL-7B & 84.0 & 84.2 & \textbf{84.5} & \textbf{84.5} & 84.4 & 84.3 & 84.1 \\
			\bottomrule[1.2pt]
		\end{tabular}
	}
\end{table*}


\section{More Visualization Results}

\begin{figure*}[!t]
    \centering
    \includegraphics[width=0.9\textwidth]{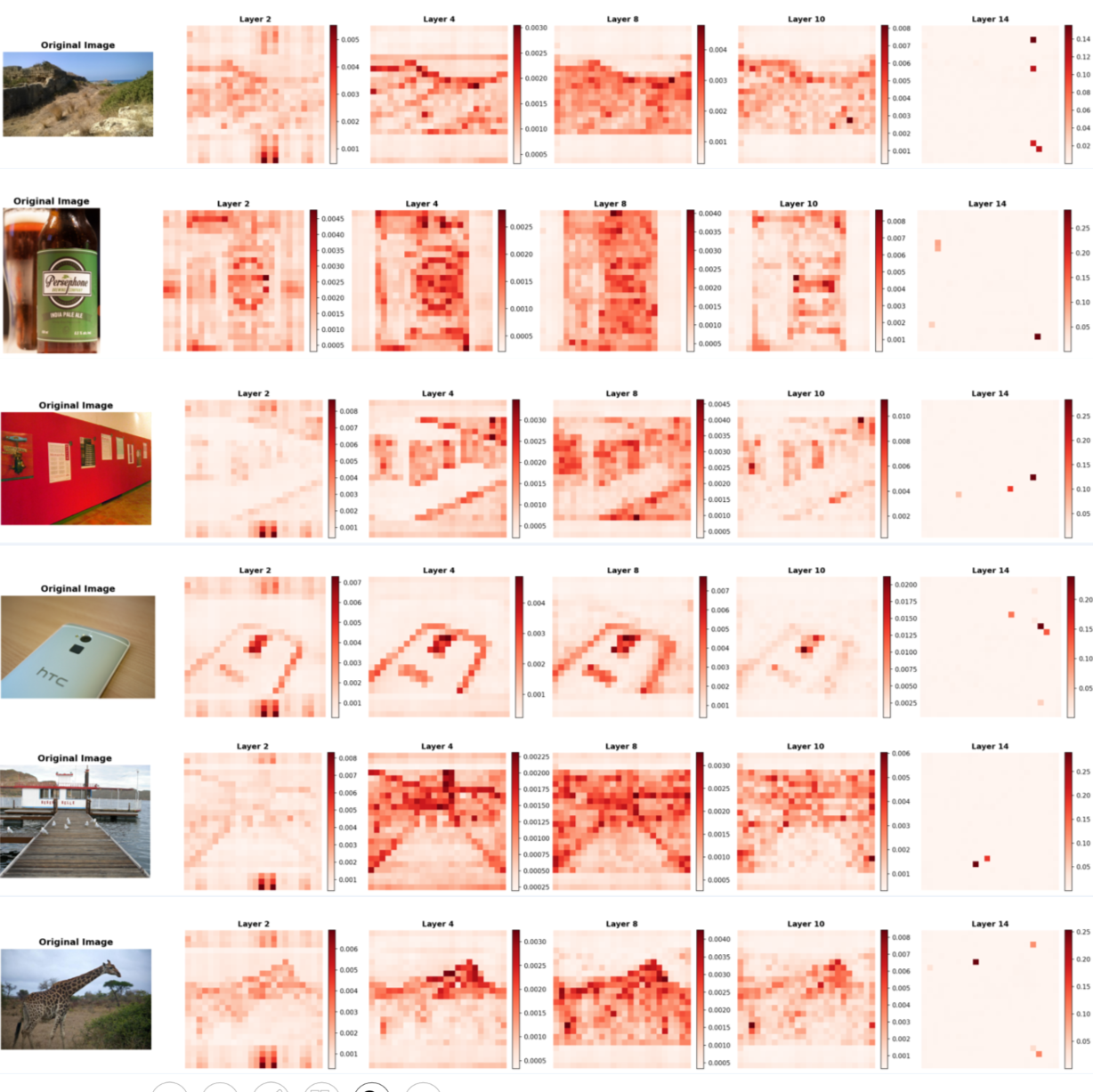} 
    \caption{\textbf{Additional visualization of attention maps.} We display CLS token attention heatmaps for six diverse samples across Layers 2, 4, 8, 10, and 14. The progression reveals a consistent pattern: shallow layers focus on \textbf{edge contours} with noisy distributions; middle layers achieve \textbf{semantic subject localization}; and deep layers converge attention onto \textbf{hub tokens} for abstract aggregation.}
    \label{fig:supp_attn_maps}
\end{figure*}

To further examine our empirical hypothesis about attention evolution, we provide additional visualizations of CLS-token attention maps across six representative samples in Fig.~\ref{fig:supp_attn_maps}. Consistent with our quantitative evidence, these examples illustrate the recurring depth-dependent pattern. In the shallow layers (Layers 2 and 4), the attention distributions are inherently \textbf{chaotic yet edge-centric}, predominantly highlighting high-frequency structural contours and object boundaries rather than complete semantic entities, consistent with strong \textbf{edge alignment} during early processing. As the network deepens to intermediate stages (Layers 8 and 10), the attention patterns transition sharply to focus on \textbf{coherent semantic subjects}, demonstrating robust localization of main objects with suppressed background noise. Finally, in the deep layers (Layer 14), we observe a pronounced convergence where attention collapses onto a few \textbf{hub tokens}, indicating the shift towards abstract semantic aggregation. These examples qualitatively support, but do not establish causally, the edge-to-object-to-aggregation hypothesis motivating our stage-specific pruning design.

\begin{figure*}[!t]
    \centering
    \includegraphics[width=\textwidth]{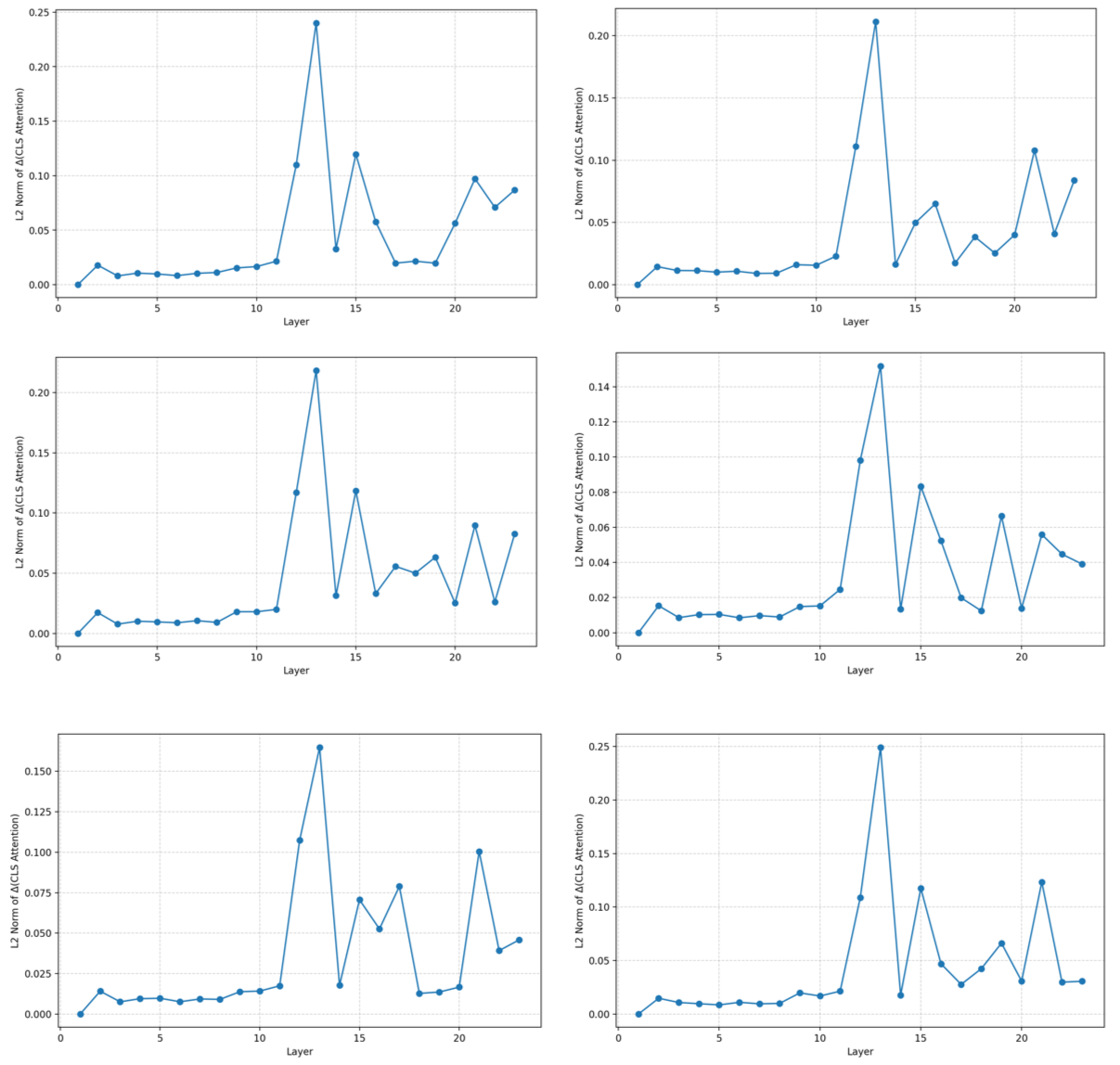}
    \caption{\textbf{Visualization of inter-layer semantic changes.} We plot the $L_2$ norm of attention differences between adjacent layers for six samples. The results consistently show \textbf{stable, low-magnitude changes} during the shallow and intermediate perception stages, contrasting sharply with the \textbf{significant instability and fluctuations} observed in the deep semantic aggregation stages. This pattern is consistent with intermittent semantic evolution in deep layers and motivates stability-aware pruning.}
    \label{fig:supp_cls_change}
\end{figure*}

Furthermore, we extend our analysis of semantic stability by visualizing the inter-layer semantic changes ($\Delta$ CLS Attention) for six additional cases in Fig.~\ref{fig:supp_cls_change}. These plots quantify the $L_2$ norm of differences between adjacent layers, reinforcing our observation of the \textbf{instability} inherent in the deep semantic aggregation phase. In the early perception stages (covering both shallow and intermediate layers), the curves remain consistently low and stable, indicating that the model's representation evolves gradually with minimal abrupt changes, suggesting high redundancy suitable for pruning. In stark contrast, once the model enters the deep semantic aggregation phase, the curves exhibit significant \textbf{fluctuations and spikes}. These peaks correspond to layers where rapid semantic evolution and critical information refinement occur. This visual evidence strongly supports our hypothesis that pruning during these unstable deep layers disrupts essential semantic induction, whereas pruning during the stable phases is safe. Consequently, a stability-adaptive strategy is crucial to safeguard these critical refinement steps while maximizing compression efficiency.

\clearpage
\end{document}